# Enhancing space–time video super-resolution via spatial–temporal feature interaction

Zijie Yue [a], Miaojing Shi [a,b,*]

[a] College of Electronic and Information Engineering, Tongji University, China
[b] Shanghai Institute of Intelligent Science and Technology, Tongji University, China

A B S T R A C T

The target of space–time video super-resolution (STVSR) is to increase both the frame rate (also referred to as the temporal resolution) and the spatial resolution of a given video. Recent approaches solve STVSR using end-to-end deep neural networks. A popular solution is to first increase the frame rate of the video; then perform feature refinement among different frame features; and at last, increase the spatial resolutions of these features. The temporal correlation among features of different frames is carefully exploited in this process. The spatial correlation among features of different (spatial) resolutions, despite being also very important, is however not emphasized. In this paper, we propose a spatial–temporal feature interaction network to enhance STVSR by exploiting both spatial and temporal correlations among features of different frames and spatial resolutions. Specifically, the spatial–temporal frame interpolation module is introduced to interpolate low- and high-resolution intermediate frame features simultaneously and interactively. The spatial–temporal local and global refinement modules are respectively deployed afterwards to exploit the spatial–temporal correlation among different features for their refinement. Finally, a novel motion consistency loss is employed to enhance the motion continuity among reconstructed frames. We conduct experiments on three standard benchmarks, Vid4, Vimeo-90K and Adobe240, and the results demonstrate that our method improves the state-of-the-art methods by a considerable margin. Our codes will be available at https://github.com/yuezijie/STINet-Space-time-Video-Super-resolution.

## 1. Introduction

The development of high-definition display techniques, *e.g.* quantum high dynamic range processor, has enabled the ultra high definition television (UHDTV) which can display videos up to 8K spatial resolution and 240 frames per second (fps) frame rate (also referred to as the temporal resolution of the video). Yet, commonly used video capturing equipment, such as smartphones and web cameras, can only capture full high definition (FHD) videos with a spatial resolution of 2K and a frame rate less than 60 fps (Kang, Jo, Oh, Vajda, & Kim, 2020). To adapt FHD videos to the UHDTV display for satisfactory visual experience, one promising way is to perform the space–time video super-resolution (STVSR) (Shechtman, Caspi, & Irani, 2002). STVSR aims to reconstruct videos to their higher spatial and temporal resolution versions. Owing to the clear motion dynamics and fine object details in reconstructed frames, STVSR is beneficial to many applications and has become a research hotspot nowadays.

Traditional STVSR approaches (Mudenagudi, Banerjee, & Kalra, 2011; Shahar, Faktor, & Irani, 2011; Shechtman & Caspi, 2005) rely on scene-oriented assumptions (*e.g.* illumination invariance (Mudenagudi et al., 2011)) to align features between given and interpolated frames. This limits their generalizability for complex space–time scenarios (Xiang et al., 2020).

With the advent of deep learning, modern STVSR has been realized as a cascade of video frame interpolation (Bao et al., 2019; Park, Kim, Lee, & Ro, 2021; Shen et al., 2021) and video super-resolution (Lai et al., 2020; Song et al., 2021; Yue et al., 2021) approaches. Despite their success, these two-stage solutions have drawbacks: the intrinsic relations between spatial and temporal information of video frames are not exploited; the computational cost is also expensive (Xiang et al., 2020; Xu et al., 2021).

To address the above issues, several one-stage STVSR methods have recently been proposed and achieved state-of-the-art performance (Cao et al., 2022; Chen, Chen, Lin, & Peng, 2023a; Chen, Yue, & Shi, 2023b; Geng et al., 2022; Haris, Shakhnarovich, & Ukita, 2020; Hu, Jiang, Nie, Zhou & and Wang, 2023; Hu, Jiang, Wang, Bai, & Hu, 2023; Huang et al., 2024; Kang et al., 2020; Shi et al., 2021; Wang et al., 2023;

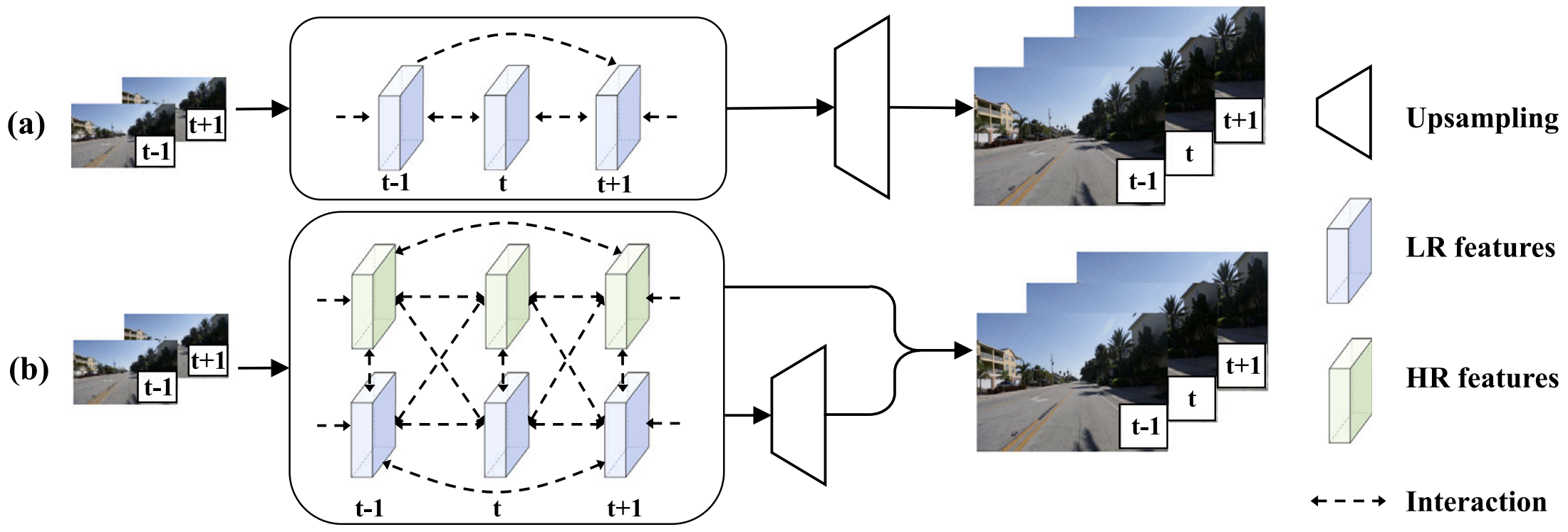

**Fig. 1.** (a) Many advanced STVSR approaches (Cao et al., 2022; Geng, Liang, Ding, & Zharkov, 2022; Kang et al., 2020; Wang, Xiang, Tian, Yang, & Liao, 2023; Xiang et al., 2020; Xu et al., 2021) begin with LR features of frames and focus on exploiting their temporal correlation in the network design. (b) Our STINet instead performs a comprehensive spatial–temporal interaction of features with different spatial resolutions (LR and HR) and temporal stamps in STVSR.

Xiang et al., 2020; Xu et al., 2021). Among them, a popular solution (Fig. 1(a)) is to first increase the frame rate of the video, then apply feature refinement among different low-resolution (LR) frame features, and finally increase the spatial resolutions of these features.

The whole process is end-to-end optimized. The temporal correlation among different frames, which concerns the information about object deformations and movements along the temporal dimension, is advantageously utilized to synthesize continuous temporal trajectories (Cao et al., 2022; Geng et al., 2022; Kang et al., 2020; Shi et al., 2021; Wang et al., 2023; Xiang et al., 2020; Xu et al., 2021). In contrast, the spatial correlation among features of different spatial resolutions is often ignored in these methods, consequently, limiting their performance in complex space–time scenarios.

We argue that exploiting temporal and spatial correlations among features of different frames and spatial resolutions in STVSR are both important due to their complementary nature for obtaining robust spatial–temporal representations (Haris et al., 2020; Hu, Jiang, Wang et al., 2023). Specifically, on one hand, the motion information encoded in frame features of different spatial resolutions is complementary (Haris et al., 2020; Hu, Jiang, Wang et al., 2023), thus aggregating spatial predictions in different resolutions can help recover finer temporal trajectories. On the other hand, consecutive frames typically exhibit similar object appearances (spatial information such as the texture patterns and structures) (Haris et al., 2020; Hu, Jiang, Wang et al., 2023), thus exploiting the correlation among temporal frame features can be beneficial for the aggregation of spatial contexts. Therefore, our primary motivation is to improve STVSR by enabling both intra-frame LR-HR (spatial) interactions and inter-frame temporal interactions simultaneously (Fig. 1(b)).

We introduce a new end-to-end spatial–temporal feature interaction network (STINet) to enhance STVSR. Given the input of a low-frame-rate low-resolution video, we first initialize its corresponding LR and HR frame features. To interpolate intermediate frame features from them, we design a new spatial–temporal frame interpolation (ST-FI) module consisting of a motion subnet and an interpolation subnet. Shareable blocks are specifically injected into the ST-FI module to facilitate the information interaction between LR and HR features.

Having the initialized and interpolated LR and HR features, on one hand, we directly optimize them with their corresponding video ground truth to distinguish between LR and HR predictions for a certain frame; on the other hand, we pass them through several spatial–temporal local refinement (ST-LR) modules, each learns two deformable sampling functions for refining the LR or HR feature with their local neighbors. An interaction subnet is proposed in the ST-LR module to enable the interaction between the predicted sampling offsets in the two functions. Afterwards, we present a spatial–temporal global refinement (ST-GR) module based on a graph convolutional network (GCN). The graph topology and edge attributes are carefully designed to enable the spatial–temporal information passing and exchanging over the whole LR and HR feature sequence. In the end, we fuse the LR and HR features per frame to obtain the high-frame-rate high-resolution video. For the network optimization, we introduce a new motion consistency loss: it computes predicted optical flows across reconstructed frames and defines absolute and relative loss terms to enforce the predictions to conform to both the ground truth and the motion continuity between frames.

In summary, the main contribution of this paper is below:

1. We introduce a novel end-to-end architecture to explicitly perform spatial–emporal feature interaction in STVSR.
2. Three key modules are highlighted: the spatial—temporal frame interpolation module, the spatial–temporal local refinement module, and the spatial–temporal global refinement module. They are carefully designed for comprehensive spatial–temporal feature interaction.
3. A new motion consistency loss is designed to integrate the motion factor into the network optimization.
4. Our method significantly outperforms the state-of-the-art approaches on standard benchmarks Vid4 (Liu & Sun, 2011), Vimeo-90K (Xue, Chen, Wu, Wei, & Freeman, 2019) and Adobe240 (Su et al., 2017).

Besides above, we notice that most state-of-the-art STVSR solutions can only support single frame interpolation between two adjacent frames whereas in reality we would need multi-frame interpolation to display a FHD video on an UHDTV. To achieve this, we enable our STINet with controllable frame interpolation rate.

## 2. Related work

### 2.1. Video frame interpolation

Video frame interpolation (VFI) synthesizes intermediate frames between temporally neighboring frames in the same level of spatial resolution. It can be mainly implemented as kernel-based (Lee et al., 2020; Shi, Liu, Shi, Dai, & Chen, 2022) or flow-based (Huang, Zhang, Heng, Shi, & Zhou, 2021; Jiang et al., 2018; Niklaus & Liu, 2018; Yan, Tan, Lin, & Shen, 2021) approaches. Kernel-based approaches generate intermediate frames by designing special convolutional kernels in deep neural networks to convolve on adjacent frames. For instance, Lee et al. (Lee et al., 2020) proposed the spatially-adaptive convolution kernel to map pixels in given frames to proper locations in the intermediate frame. Shi et al. (Shi et al., 2022) introduced the generalized deformable convolution kernels to select existing pixels for synthesizing new pixels. Flow-based approaches predict flows to map pixels from adjacent frames to new locations in the intermediate frame. For instance, Niklaus et al. (Niklaus & Liu, 2018) estimated bidirectional optical flows between given frames and passed them with corresponding pixel-level context maps to a frame synthesis network. To reduce artifacts around boundaries of moving objects in intermediate frames, Jiang et al. (Jiang et al., 2018) employed two consecutive U-Nets to refine the estimated optical flows before feeding them into the frame synthesis network. Huang et al. (Huang et al., 2021) designed a knowledge distillation model which constrains flows predicted from the student model to be consistent with that from the teacher model.

### 2.2. Video super-resolution

Video super-resolution (VSR) is to reconstruct HR frames from their LR counterparts. Existing approaches perform VSR in two main steps (Haris, Shakhnarovich, & Ukita, 2019; Tao, Gao, Liao, Wang, & Jia, 2017; Tu, Li, Xie, Liu, Zhang, Li, & Yuan, 2022; Xue et al., 2019; Yi, Wang, Jiang, Shao, & Ma, 2020): first, aggregating spatial information from the reference and its neighboring LR frames; second, performing feature alignment and warping to reconstruct the corresponding HR frame. Haris et al. (Haris et al., 2019) iteratively calculated the residual maps to capture spatial context differences between the reference and neighboring LR frames, then back-projected these residuals on the reference frame to produce its HR version. Jo et al. (Jo, Oh, Kang, & Kim, 2018) predicted dynamics upsampling filters via a dynamic filter generation network, which are then used to filter LR frames for enlarging their spatial resolutions. Zhang et al. (Zhang, Shao, Liang, Liu, & Shen, 2021) designed a multi-branch network to learn multi-scale spatial features from LR frames to boost the VSR performance. Other methods have utilized the convolutional LSTM (Huang, Wang, & Wang, 2018), generative adversarial network (Lucas, Lopez-Tapia, Molina, & Katsaggelos, 2019), and graph neural network (Tarasiewicz, Nalepa, & Kawulok, 2021) to reconstruct HR frames.

### 2.3. Space–time video super-resolution

Space–time video super-resolution (STVSR) aims to increase both the spatial and temporal resolutions of videos. Many traditional approaches rely on scene-oriented assumptions to assist the feature alignment between adjacent LR frames before reconstructing HR frames (Mudenagudi et al., 2011; Shahar et al., 2011; Shechtman & Caspi, 2005). For example, Mudenagudi et al. (Mudenagudi et al., 2011) assumed no illumination changes among neighboring frames and employed the graph-cut optimization to reconstruct high-frame-rate HR videos from low-frame-rate LR videos.

Recently, deep learning models have demonstrated robust feature alignment and refinement capabilities in STVSR (Cao et al., 2022; Chen et al., 2023a; Geng et al., 2022; Haris et al., 2020; Hu, Jiang, Nie et al., 2023; Hu, Jiang, Wang et al., 2023; Huang et al., 2024; Kang et al., 2020; Shi et al., 2021; Wang et al., 2023; Xiang et al., 2020; Xu et al., 2021). Xiang et al. (Xiang et al., 2020) designed the deformable feature alignment functions to perform intermediate frame interpolation. Xu et al. (Xu et al., 2021) enabled the frame reconstruction at an arbitrary intermediate moment. Shi et al. (Shi et al., 2021) developed the generalized Pixelshuffle layer to make the spatial resolution of reconstructed videos controllable.

Geng et al. (Geng et al., 2022) designed a cascaded U-Net style structure to achieve comparable performance with state of the art in STVSR yet using lesser network parameters. Cao et al. (Cao et al., 2022) proposed a Fourier data transform layer and a recurrent video enhancement layer to respectively solve the motion blur and motion aliasing problems in the reconstructed videos. Chen et al. (2023a) explicitly encoded forward motion information as input to the model, aiding in recovering object motion trajectories. Huang et al. (2024) implemented scale-adaptive feature propagation and fusion for efficient STVSR. Wang et al. (2023) devised a long–short term feature interpolation module to aggregate visual content from multiple neighboring frames for intermediate frame reconstruction. Hu, Jiang, Nie et al. (2023) designed both a backward and a forward recurrent module to model the long-range spatial–temporal correlations for STVSR.

### 2.4. Comparison of ours to state of the art

The designing spirit of our STINet is to enhancing spatial–temporal feature interaction among features of different frames and spatial resolutions to improve STVSR, which has not been explicitly exploited before in the STVSR task. As mentioned in Section 1, many state-of-the-art approaches (Cao et al., 2022; Geng et al., 2022; Kang et al., 2020; Shi et al., 2021; Wang et al., 2023; Xiang et al., 2020; Xu et al., 2021) begin with LR features of frames and focus on exploiting their temporal correlation. Haris et al. (2020) has tried to apply local refinement from HR features to LR features or vice versa but there is no mutual information exchange in this process. In contrast, we explicitly perform spatial–temporal feature interaction in STVSR, where HR and LR features are comprehensively interacted throughout the network including feature interpolation, local feature refinement between adjacent frames and global feature refinement across the whole video sequence.

For the feature interpolation, unlike previous work (Haris et al., 2020) which interpolates LR and HR features independently, we design a novel ST-FI module with shareable blocks to enable interaction between them. For the local feature refinement, Haris et al. (2020), Shi et al. (2021) leverage residual blocks to compute residual maps between features of adjacent frames for their refinement. We instead refine certain frame features via DCN with motion offsets learned from each frame's neighbors. Cao et al. (2022), Xiang et al. (2020), Xu et al. (2021) also utilize DCN for local feature refinement, but they only learn one set of motion offsets from adjacent LR features. Unlike them, our ST-LR learns two sets of motion offsets with different spatial resolutions and we introduce an interaction subnet to enable their interaction. For the global feature refinement, (Cao et al., 2022; Haris et al., 2020; Shi et al., 2021) do not have such a phase. Xiang et al. (2020), Xu et al. (2021) utilize the BDConvLSTM to exploit the temporal correlation over the sequence of LR features. Our ST-GR module builds upon the GCN to exploit spatial–temporal correlation over the whole sequence of LR and HR features. We carefully design the graph topology and three novel edge attributes to enhance the exchange of complementary spatial–temporal information among feature sequences. Besides these three modules, we also propose a novel optical flow adaption method and a novel motion consistency loss (MCL) to help synthesize continuous motion trajectory.

## 3. Method

### 3.1. Overview

The overview of our STINet is illustrated in Fig. 2. Our goal is to reconstruct the corresponding high-frame-rate HR video sequence $\mathcal{V}$ from the input of low-frame-rate LR video sequence $\mathcal{F}$. STINet contains four phases: spatial–temporal feature interpolation (ST-FI), spatial–temporal local and global refinements (ST-LR and ST-GR), and finally HR frames reconstruction. ST-FI produces LR and HR features of frames, which enables both the spatial (*e.g.* texture patterns and structures of objects) and temporal (*e.g.* deformations and movements of objects) information interaction among features of different resolutions and frames. ST-LR and ST-GR elicit the spatial–temporal correlation of features for their local and global refinement, respectively.

Specifically, given two adjacent LR frames $F_{t-1}$ and $F_{t+1}$, we can calculate their bidirectional optical flows $O_{t-1\to t+1}$ and $O_{t+1\to t-1}$. We first initialize the LR and HR features, *i.e.* $L_{t-1}$ and $H_{t-1}$, $L_{t+1}$ and $H_{t+1}$, for $F_{t-1}$ and $F_{t+1}$, respectively. The initialization is through the spatial super-resolution (S-SR) module (Haris, Shakhnarovich, & Ukita, 2018), which for instance takes the input of $F_{t-1}$ to output $H_{t-1}$ and $L_{t-1}$. To synthesize intermediate frame features, we adapt $O_{t-1\to t+1}$ and $O_{t+1\to t-1}$ to obtain optical flows from the target intermediate frame to adjacent frames. Using adapted optical flows as input together with $L_{t-1}$ and $L_{t+1}$ or $H_{t-1}$ and $H_{t+1}$, we interpolate intermediate LR and HR frame features through a spatial–temporal frame interpolation (ST-FI) module (Section 3.2). Having LR and HR feature sequences, *i.e.* $\mathcal{L} = \{L_t\}$, $\mathcal{H} = \{H_t\}$, on one hand, we optimize them with the corresponding video ground truth *i.e.* $\mathcal{F}^{gt}$ and $\mathcal{V}^{gt}$, to enforce them to learn the discriminative information; on the other hand, we pass them through several shareable spatial–temporal local refinement (ST-LR) modules to refine each frame feature by its neighbors (Section 3.3). Afterwards, we feed the locally-refined features $\mathcal{L}^R$ and $\mathcal{H}^R$ to a spatial–temporal global refinement (ST-GR) module to further refine them over the whole sequence (Section 3.4), output denoted by $\mathcal{L}^G$ and $\mathcal{H}^G$. In the end, we project $\mathcal{L}^G$ via two Pixelshuffle layers (Shi et al., 2016) followed by a 1 × 1 convolutional layer and project $\mathcal{H}^G$ directly by a 1 × 1 convolutional layer to obtain the corresponding video frames $\mathcal{V}^L$ and $\mathcal{V}^H$, respectively. $\mathcal{V}^L$ and $\mathcal{V}^H$ are added to produce the final HR frames $\mathcal{V}$ (Section 3.5).

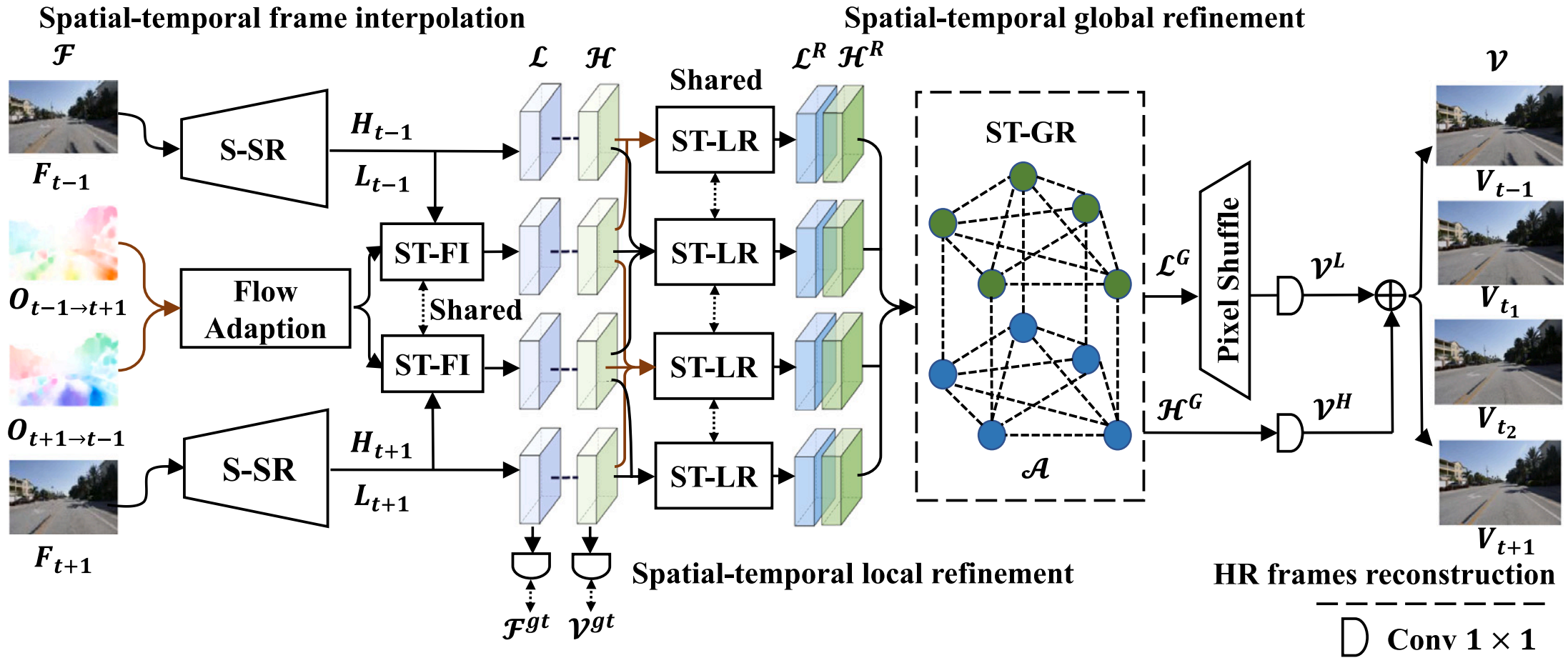

**Fig. 2.** STINet contains four main phases. Phase 1: we initialize the LR and HR features of given frames and interpolate intermediate frame features in the proposed ST-FI module. Phase 2: we pass the LR and HR features into a number of shareable ST-LR modules to capture motion cues from neighboring frames for local feature refinement. Phase 3: we construct a graph network in the ST-GR module which passes complementary spatial–temporal information over whole feature sequence for global feature refinement. Phase 4: we fuse LR and HR features to reconstruct high-frame-rate HR video sequence $\mathcal{V}$.

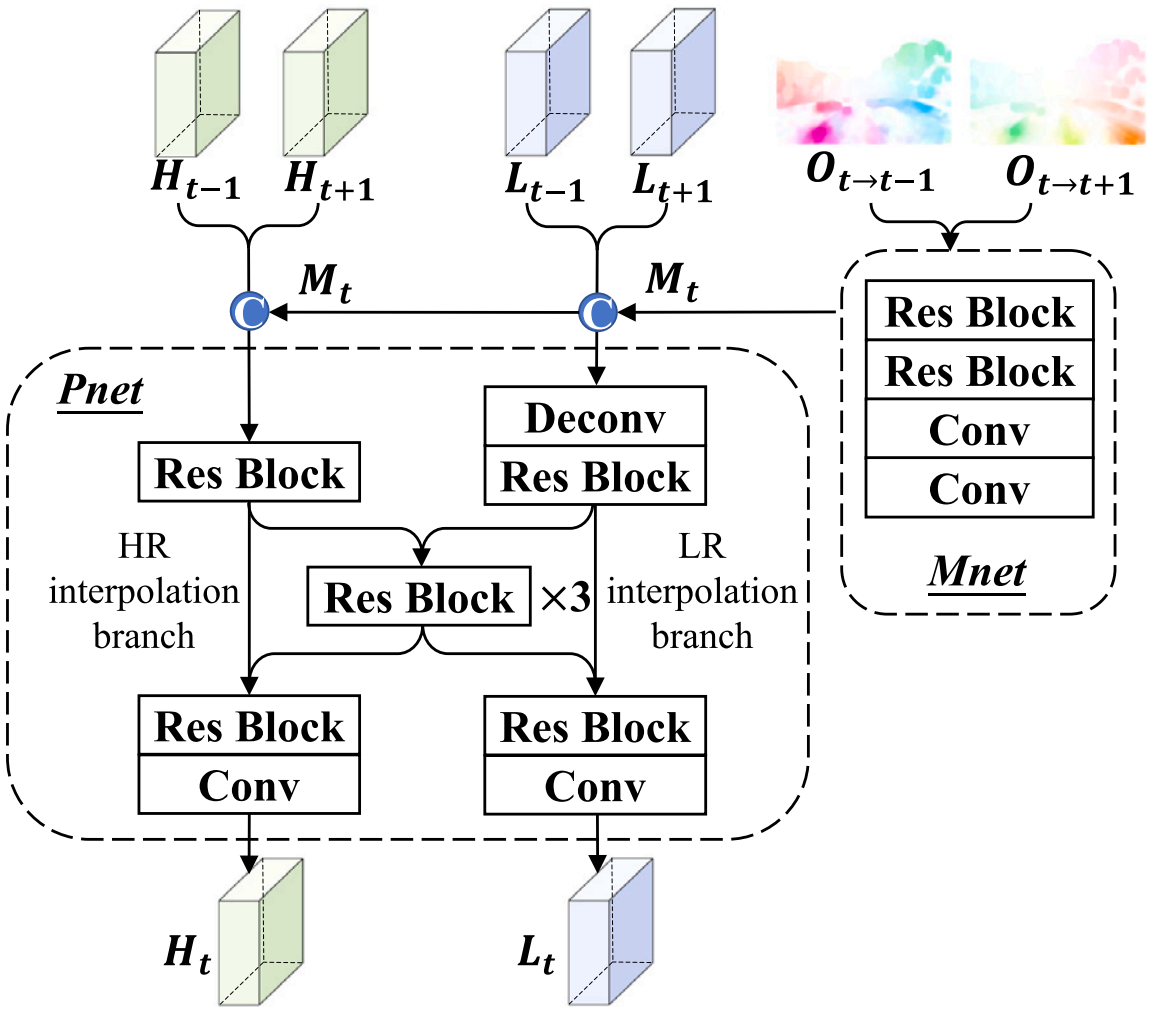

**Fig. 3.** The ST-FI module interpolates intermediate features ($L_t$, $H_t$) based on both optical flows ($O_{t\to t-1}$, $O_{t\to t+1}$) and initialized LR and HR features ($L_{t-1}$, $L_{t+1}$, $H_{t-1}$, $H_{t+1}$).

Our network supports multi-frame interpolation given two adjacent frames ($F_{t-1}$ and $F_{t+1}$), as illustrated by Fig. 2. Without loss of generality, below we present it for interpolating a single frame ($t$th frame $V_t$).

### 3.2. Spatial–temporal feature interpolation

At this phase, we first introduce an optical flow adaption method to obtain optical flows from each intermediate frame to adjacent frames; next, since LR and HR features are complementary (Zhang, Chen, Lin, Chandran, & Jing, 2021), we design the ST-FI module with a novel interpolation subnet to perform LR and HR feature interaction during the intermediate feature interpolation.

First, in order to interpolate intermediate frame features from features of adjacent frames, we follow flow-based methods (Jiang et al., 2018; Niklaus & Liu, 2018; Shi et al., 2021) to rely on optical flows which describe the velocity distribution of object movement between frames. Different from them, we propose a simple yet effective way to obtain optical flows between the target intermediate frame and their adjacent frames. Having the flows between $F_{t-1}$ and $F_{t+1}$, *i.e.* $O_{t-1\to t+1}$ and $O_{t+1\to t-1}$, to adapt them to represent motion information from the frame of an arbitrary moment $t' \in (t-1, t+1)$ to that of moment $t-1/t+1$, we assume the motion change within the short period $(t+1-(t-1)=2)$ is linear, such that:

$$O_{t'\to t-1} = \frac{t'-(t-1)}{2} \times O_{t+1\to t-1}$$
$$O_{t'\to t+1} = \frac{(t+1)-t'}{2} \times O_{t-1\to t+1} \tag{1}$$

This equation can be used for the multi-frame interpolation. For simplicity, here we use $t' = t$, so $O_{t\to t-1} = 0.5 \times O_{t+1\to t-1}$, $O_{t\to t+1} = 0.5 \times O_{t-1\to t+1}$.

Next, we perform LR and HR intermediate feature interpolation via a spatial–temporal feature interpolation (ST-FI) module. Its structure is shown in Fig. 3, which consists of a motion subnet (Mnet) and an interpolation subnet (Pnet). The Mnet is inspired by Haris et al. (2020) to model a motion representation $M_t$ from $O_{t\to t-1}$ and $O_{t\to t+1}$, which can help spatial alignment from given features to intermediate ones. It is composed of two residual blocks and two convolutional layers.

The Pnet is designed to take the input of $M_t$, $L_{t-1}$, $L_{t+1}$, $H_{t-1}$ and $H_{t+1}$ to synthesize intermediate frame features $L_t$ and $H_t$: We concatenate $M_t$ with $L_{t-1}$ ($H_{t-1}$) and $L_{t+1}$ ($H_{t-1}$) as an input for the LR (HR) interpolation branch, respectively. The two branches are jointed through three shareable residual blocks, which help encode mutual and complementary contexts among frame features of different spatial–temporal information. Skip connection is applied to concatenate the input and output of these shareable blocks for LR or HR interpolation branch.

The output feature sequences, $\mathcal{L} = \{L_t\}$ and $\mathcal{H} = \{H_t\}$ for LR and HR features of both intermediate and adjacent frames, are passed through separated 1 × 1 convs and supervised by the corresponding ground truth, $\mathcal{F}^{gt}$ and $\mathcal{V}^{gt}$.

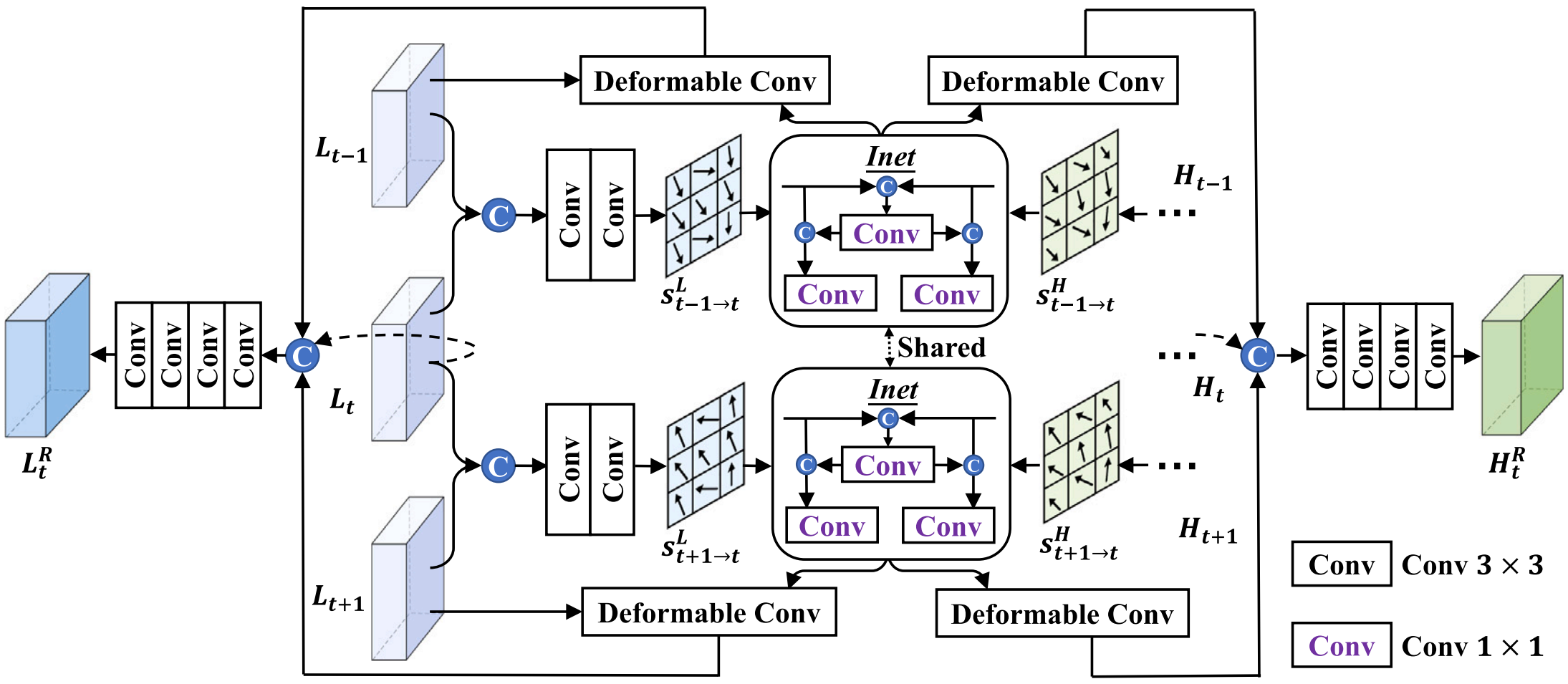

**Fig. 4.** An example architecture of the ST-LR module which takes the input of $L_{t-1}$, $L_t$, and $L_{t+1}$ from the left side, and $H_{t-1}$, $H_t$, and $H_{t+1}$ from the right side to refine $L_t$ and $H_t$.

### 3.3. Spatial–temporal local refinement

This phase aims to refine LR and HR features of each frame by extracting short-term motion cues from their temporal neighbors and learning deformable sampling functions (Xiang et al., 2020; Xu et al., 2021). Specifically, we learn one set of sampling offsets for each LR or HR feature with its certain neighbor, respectively, and enable the information exchange between the learned two sets of offsets via a new interaction subnet.

Fig. 4 illustrates an example of ST-LR that refines $L_t(H_t)$ using $L_{t-1}(H_{t-1})$ and $L_{t+1}(H_{t+1})$. Having a look at the left side, we concatenate $L_t$ with $L_{t-1}$ and $L_{t+1}$ respectively and utilize two convolutional layers each to predict the offsets $s^L_{t-1\to t}$ and $s^L_{t+1\to t}$. These offsets implicitly describe the forward and backward motion cues from $L_{t-1}$ and $L_{t+1}$ to $L_t$. Similar offsets $s^H_{t-1\to t}$ and $s^H_{t+1\to t}$ can be estimated from HR features $\{H_{t-1}, H_t, H_{t+1}\}$ in the right side. The corresponding offsets pairs, $s^L_{t-1\to t}$ and $s^H_{t-1\to t}$, $s^L_{t+1\to t}$ and $s^H_{t+1\to t}$, are of different resolutions and are fed into a shareable interaction subnet (Inet, specified below) for information exchange. The refined offsets are served as the kernel parameters for deformable convolutional networks (DCN) (Dai et al., 2017), which are used to align $L_{t-1}/L_{t+1}$ towards $L_t$, and $H_{t-1}/H_{t+1}$ towards $H_t$. Last, we respectively concatenate $L_t$ and $H_t$ with the aligned feature maps and apply four convolutional layers each to obtain locally-refined features $L^R_t$ and $H^R_t$.

We specify our Inet by taking an example of $s^L_{t-1\to t}$ and $s^H_{t-1\to t}$. They are first concatenated and passed through a convolutional layer to capture their complementary information regarding the short-term object motions between $t-1$ and $t$th frames. The output is respectively concatenated with the original input $s^L_{t-1\to t}$ and $s^H_{t-1\to t}$ followed by one convolutional layer each to produce the refined offsets, $\hat{s}^L_{t-1\to t}$ and $\hat{s}^H_{t-1\to t}$. This can be understood as a modulation process on the input $s^L_{t-1\to t}$ and $s^H_{t-1\to t}$, which enhances their salient information while suppressing their trivial information. Backward offsets $\hat{s}^L_{t+1\to t}$ and $\hat{s}^H_{t+1\to t}$ can be obtained similarly.

The ST-LR module is applied to each feature in $\mathcal{L}$ and $\mathcal{H}$ with a sliding window of size 3. For the first (last) feature in the sequence, its previous (next) neighbor is itself. By the end of this phase, we have refined features $\mathcal{L}^R$ and $\mathcal{H}^R$.

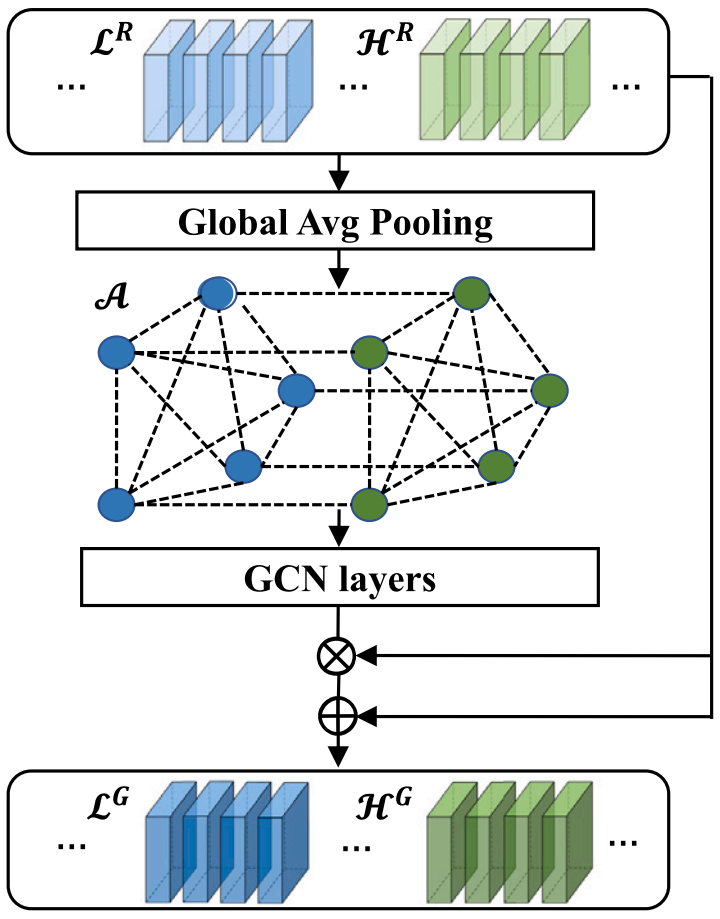

**Fig. 5.** The ST-GR module performs globally spatial–temporal information passing and exchanging over the whole feature sequence via the graph convolutional network (GCN).

### 3.4. Spatial–temporal global refinement

This phase aims to refine features of each frame over the whole sequence.

Since the graph convolutional network (GCN) has been proved effective in video matting (Wang, Liu, Tian, Li, & Yang, 2021) and object tracking (Dai et al., 2021) tasks for its powerful information aggregation ability, we build the ST-GR module upon the GCN. We introduce a novel way to define the edges and edge attributes among different frame features based on their spatial–temporal correlation. It can thus globally exchange the complementary information regarding object spatial structures and temporal movements among LR and HR features of different frames. The graph topology is particularly suitable for this purpose. This cannot be achieved by the BDConvLSTMas used in Xiang et al. (2020), Xu et al. (2021). BDConvLSTM is good at encoding temporal correlation among frames but not spatial correlation among features with different resolutions. Below, we specify our module, its structure is illustrated in Fig. 5.

Features in $\mathcal{L}^R$ and $\mathcal{H}^R$ are of dimension $w \times h \times c$. They are firstly fed into a global-average pooling layer to obtain the frame-level global features of dimension $1 \times 1 \times c$, which are used as node features to construct the graph $\mathcal{A}$. The structure of $\mathcal{A}$ is defined by an adjacency matrix, which represents the connections among nodes. First, each node is connected to its corresponding HR or LR counterpart as they describe the same frame of different spatial resolutions. Second, each node is connected to other nodes of the same spatial-resolution level yet different frames. The connections enable both spatial and temporal correlations between features.

Next, for each connection (edge), we define the edge attribute as a triplet $E = (E^F, E^P, E^T)$ of three factors: (1) Frame-level similarity $E^F$ measures the cosine similarity between the frame-level global features (node features) of two connected nodes; (2) Pixel-level similarity $E^P$ averages the cosine similarities between pixel-level features (pixel-wise features taken from the $w \times h \times c$ tensor of $L^R/H^R$) of two connected nodes. For the calculation between features with different spatial resolutions, we downsample HR features to match the resolution of LR ones; (3) Temporal closeness $E^T$ measures the temporal-distance of two connected nodes. Given the time stamps of $t_1$ and $t_2$ for two nodes, $E^T$ is computed as $1 - \sigma\left(|t_1 - t_2|\right)$; $\sigma$ indicates the sigmoid function. $E^F$ and $E^P$ measure the globally/locally visual similarity of two features. $E^T$ measures the temporal closeness of two frames. For any two nodes that are either visually similar or temporally close, they are assigned with bigger weights by $E^F, E^P$ or $E^T$ such that more spatial–temporal information will be exchanged through them in the graph.

Once $\mathcal{A}$ is initialized, we feed the node features together with their connections defined in $\mathcal{A}$ into four GraphSAGE layers (Hamilton, Ying, & Leskovec, 2017) to perform spatial–temporal information passing and exchanging for global refinement. Each output node feature ($1\times1\times c$) is channel-wisely multiplied to its corresponding original HR/LR feature ($w \times h \times c$); the resulting feature serves as a modulation term and is added back to the original feature for refinement. We denote the output of this module by $\mathcal{L}^G = \left\{L_t^G\right\}$ and $\mathcal{H}^G = \left\{H_t^G\right\}$ for LR and HR feature sequences.

### 3.5. High-frame-rate high resolution video reconstruction

We follow Xu et al. (2021) to feed $\mathcal{L}^G$ into two Pixelshuffle layers for spatial super-resolution. They are followed by a $1 \times 1$ convolutional layer to project the features to 2-dimensional images $\mathcal{V}^L$. $\mathcal{H}^G$ is projected via another $1 \times 1$ convolutional layer to obtain $\mathcal{V}^H$. They are added to obtain the high-frame-rate HR video $\mathcal{V} = \left\{V_t\right\}$: $V_t = V_t^L + V_t^H$.

### 3.6. Network optimization

Three loss functions are leveraged to optimize the network parameters: the reconstruction loss, the perceptual loss, and the motion consistency loss. The former two are commonly used in STVSR approaches (Geng et al., 2022; Haris et al., 2020; Shi et al., 2021; Xiang et al., 2020; Xu et al., 2021): $l_{rec} = \sum_{t=1}^{N} \left\| V_t - V_t^{gt} \right\|_2^2$; $l_{per} = \sum_{t=1}^{N} \left\| \phi(V_t) - \phi(V_t^{gt}) \right\|_2^2$, where $V_t$ and $V_t^{gt}$ signifies the prediction and ground truth of $t$th HR frame; $N$ is the total number of frames; $\phi(\cdot)$ is the feature extractor of the pretrained VGG-19 (Simonyan & Zisserman, 2015).

**Motion consistency loss (MCL).** $l_{rec}$ and $l_{per}$ measures the pixel-level and frame-level visual differences between the prediction and ground truth. We propose a novel motion consistency loss to integrate the temporal factor into the optimization, which can help reconstruct motions between frames more accurately. To elicit such temporal information from our predictions, we compute the optical flows between any two predicted frames, *i.e.* $O_{t\to p}^{pred}$ for HR frames $V_t$ and $V_p$. MCL is defined on the estimated optical flows consisting of two sub terms: the absolute and relative loss. For the absolute loss, we obtain the ground truth optical flows $\mathcal{O}^{gt}$ from $\mathcal{V}^{gt}$, *e.g.* $O_{t\to p}^{gt}$ for HR ground truth frames $V_t^{gt}$ and $V_p^{gt}$, and compute the mean squared error between the estimated and ground truth flow maps:

$$l_{abs} = \sum_{t=1}^{N} \sum_{p=1}^{N} \left\| O_{t\to p}^{pred} - O_{t\to p}^{gt} \right\|_2^2. \tag{2}$$

Notice (Huang et al., 2021; Zhou, Lu, Li, Yan, & Xue, 2021) have a similar flow loss to Eq. (2) but they only supervise flows between adjacent frames which can be seen as a special case of Eq. (2). Moreover, they use L1 loss while we use L2 loss as L2 loss works better for us empirically.

For the relative loss, our observation is that objects move smoothly over frames and their motions are normally accumulated towards a certain direction over a short period of time. This means, given three consecutive HR frames, $V_{t-1}$, $V_t$ and $V_{t+1}$, the motions of objects between frame $V_{t-1}$ and $V_{t+1}$ should normally be bigger than that between $V_{t-1}$ and $V_t$. In practice, we use optical flows to represent motions and define a novel loss $l_{rel}$,

$$l_{rel} = \sum_{t=2}^{N-1} \sum_{i=1}^{I} \sum_{j=1}^{2} max(sgn(O_{t-1\to t,i,j}^{pred}) \times (O_{t-1\to t,i,j}^{pred} - O_{t-1\to t+1,i,j}^{pred}), 0) \tag{3}$$

where $i, j$ represent the $i$th pixel $j$th channel of the optical flows. There are in total $I$ pixels and 2 channels ($x$ and $y$ coordinates). $sgn$ indicates the sign (motion direction) of $O_{t-1\to t,i,j}^{pred}$. The interior loss equals to zero as long as $O_{t-1\to t,i,j}^{pred}$ has the same motion direction with $O_{t-1\to t+1,i,j}^{pred}$ and $|O_{t-1\to t,i,j}^{pred}|$ is smaller than $|O_{t-1\to t+1,i,j}^{pred}|$, which agrees with the above motion constraint. We apply a rather weak constraint here to avoid over-fitting, especially since we already have a strong constraint in the absolute term. In practice, the relative term brings more benefits to the STVSR performance than the absolute term in MCL.

The overall loss function is a linear combination of the above losses:

$$l = l_{rec} + l'_{rec} + \lambda_1 \times l_{per} + \lambda_2 \times l_{mcl}, \tag{4}$$

where $l_{mcl} = l_{abs} + l_{rel}$. We use $l'_{rec}$ to represent the reconstruction loss we add at Phase 1 (Section 3.2).

Adding $l_{per}$ and $l_{mcl}$ at Phase 1 does not offer additional improvement in practice. $\lambda_1$ and $\lambda_2$ are loss weights.

## 4. Experiments

### 4.1. Datasets

The training set of this work is Vimeo-90K *septuplet* dataset (Xue et al., 2019), which is widely used in previous STVSR approaches (Geng et al., 2022; Xiang et al., 2020; Xu et al., 2021). This dataset contains 91,701 video sequences of resolution $448 \times 256$. Each sequence contains 7 HR frames. Following (Xiang et al., 2020), we downsample the odd-index frames of each video sequence to their LR counterparts of resolution $112 \times 64$ using bicubic interpolation. They are used as the training inputs to reconstruct original 7 HR frames. To evaluate our method, we follow (Geng et al., 2022; Xiang et al., 2020; Xu et al., 2021) to use the Vid4 (Liu & Sun, 2011), Vimeo-90K *septuplet test* set and Adobe240 (Su et al., 2017). The Vimeo-90K *septuplet test* set is split into three subsets corresponding to fast, medium and slow motions; there are 1225, 4977 and 1613 video sequences in each subset, respectively.

### 4.2. Implementation details and evaluation protocol

**Implementation details.** We initialize the learning rate as $1\times10^{-4}$ and decrease it by a factor of 4 for every $10^5$ iterations. The total number of iterations is $5\times10^5$ with the batch size being 30. Network parameters are initialized using Kaiming initialization (He, Zhang, Ren, & Sun, 2015) and optimized by the Adam optimizer (Kingma & Ba, 2017) with a momentum of 0.9. Moreover, following Xiang et al. (2020), the input LR frames are randomly cropped to $32 \times 32$ image patches, while their corresponding $128 \times 128$ image patches in HR frames are treated as ground truth. The training set is augmented by randomly flipping, rotating (180°) and mirroring. Loss weights $\lambda_1$ and $\lambda_2$ are both 0.1. All experiments were performed on four NVIDIA RTX 2080 GPUs.

**Table 1**
Comparison of STINet to state-of-the-art two-stage (top block) and one-stage (middle block) STVSR approaches. The speed is calculated on Vid4 (Liu & Sun, 2011). ↑ indicates that the larger the better. We indicate the best performing two-stage approaches in shadow. We highlight the best and second best performing one-stage approaches in **red** and **blue**.

| Method | Vid4 | | | Vimeo-Fast | | | Vimeo-Medium | | | Vimeo-Slow | | | Adobe240 | | | Speed | Param |
|---|---|---|---|---|---|---|---|---|---|---|---|---|---|---|---|---|---|
| VFI+VSR/STVSR | PSNR↑ | SSIM↑ | LPIPS↓ | PSNR↑ | SSIM↑ | LPIPS↓ | PSNR↑ | SSIM↑ | LPIPS↓ | PSNR↑ | SSIM↑ | LPIPS↓ | PSNR↑ | SSIM↑ | LPIPS↓ | fps↑ | Millions↓ |
| SepConv (Niklaus, Mai, & Liu, 2017)+Bicubic | 23.51 | 0.6273 | 0.3572 | 32.27 | 0.8890 | 0.1925 | 30.61 | 0.8633 | 0.1961 | 29.04 | 0.8290 | 0.2012 | 26.61 | 0.7457 | 0.2449 | – | 21.7 |
| SepConv (Niklaus et al., 2017)+RCAN (Zhang, Li et al., 2018) | 24.92 | 0.7236 | 0.3104 | 34.97 | 0.9195 | 0.1365 | 33.59 | 0.9125 | 0.1625 | 32.13 | 0.8967 | 0.1943 | 26.84 | 0.7488 | 0.2396 | 2.72 | 21.7+16.0 |
| SepConv (Niklaus et al., 2017)+RBPN (Haris et al., 2019) | 26.08 | 0.7751 | 0.2262 | 35.07 | 0.9238 | 0.1320 | 34.09 | 0.9229 | 0.1584 | 32.77 | 0.9090 | 0.1886 | 30.18 | 0.8706 | 0.2095 | 2.35 | 21.7+12.7 |
| SepConv (Niklaus et al., 2017)+EDVR (Wang, Chan, Yu, Dong, & Loy, 2019) | 25.93 | 0.7792 | 0.2310 | 35.23 | 0.9252 | 0.1284 | 34.22 | 0.9240 | 0.1583 | 32.96 | 0.9112 | 0.1851 | 30.28 | 0.8745 | 0.2120 | 6.71 | 21.7+20.7 |
| DAIN (Bao et al., 2019)+Bicubic | 23.55 | 0.6268 | 0.3585 | 32.41 | 0.8910 | 0.1843 | 30.67 | 0.8636 | 0.1971 | 29.06 | 0.8289 | 0.2183 | 26.61 | 0.7453 | 0.2475 | – | 24.0 |
| DAIN (Bao et al., 2019)+RCAN (Zhang, Li et al., 2018) | 25.03 | 0.7261 | 0.2942 | 35.27 | 0.9242 | 0.1301 | 33.82 | 0.9146 | 0.1730 | 32.26 | 0.8974 | 0.1874 | 26.86 | 0.7489 | 0.2388 | 2.63 | 24.0+16.0 |
| DAIN (Bao et al., 2019)+RBPN (Haris et al., 2019) | 25.96 | 0.7784 | 0.2279 | 35.55 | 0.9300 | 0.1273 | 34.45 | 0.9262 | 0.1498 | 32.92 | 0.9097 | 0.1806 | 30.29 | 0.8709 | 0.2224 | 2.00 | 24.0+12.7 |
| DAIN (Bao et al., 2019)+EDVR (Wang et al., 2019) | 26.12 | 0.7836 | 0.2188 | 35.81 | 0.9323 | 0.1236 | 34.66 | 0.9281 | 0.1526 | 33.11 | 0.9119 | 0.1829 | 30.40 | 0.8749 | 0.1973 | 5.48 | 24.0+20.7 |
| BMBC (Park, Ko, Lee, & Kim, 2020)+TTVSR (Liu, Yang, Fu, & Qian, 2022) | 26.14 | 0.7920 | 0.2207 | 35.98 | 0.9330 | 0.1147 | 34.67 | 0.9287 | 0.1332 | 33.14 | 0.9123 | 0.1756 | 30.32 | 0.8698 | 0.2158 | 1.63 | 11.0+6.8 |
| BMBC (Park et al., 2020)+ETDM (Isobe et al., 2022) | 26.20 | 0.7926 | 0.2174 | 36.00 | 0.9347 | 0.1110 | 34.75 | 0.9288 | 0.1309 | 33.14 | 0.9126 | 0.1730 | 30.40 | 0.8711 | 0.1991 | 1.49 | 11.0+8.4 |
| ABME (Park, Lee, & Kim, 2021) +TTVSR (Liu et al., 2022) | 26.30 | 0.7937 | 0.2136 | 36.23 | 0.9358 | 0.1089 | 35.00 | 0.9352 | 0.1327 | 33.20 | 0.9134 | 0.1747 | 30.59 | 0.8754 | 0.1962 | 3.87 | 18.1+6.8 |
| ABME (Park, Lee, & Kim, 2021)+ETDM (Isobe et al., 2022) | 26.38 | 0.7958 | 0.2091 | 36.24 | 0.9365 | 0.1088 | 35.08 | 0.9355 | 0.1313 | 33.28 | 0.9138 | 0.1723 | 30.54 | 0.8750 | 0.1977 | 3.74 | 18.1+8.4 |
| STARnet (Haris et al., 2020) | 26.06 | 0.8046 | 0.2280 | 36.19 | 0.9368 | 0.1154 | 34.86 | 0.9356 | 0.1376 | 33.10 | 0.9164 | 0.1725 | 29.92 | 0.8589 | 0.1901 | 14.15 | 111.61 |
| Zooming Slow-Mo (Xiang et al., 2020) | 26.31 | 0.7976 | 0.2127 | 36.81 | 0.9415 | 0.1022 | 35.41 | 0.9361 | 0.1303 | 33.36 | 0.9138 | 0.1693 | 30.34 | 0.8713 | 0.1760 | 16.09 | 11.10 |
| DAVSR (Cao et al., 2022) | 23.85 | 0.7460 | 0.3460 | – | – | – | – | – | – | – | – | – | – | – | – | 14.88 | 8.00 |
| TMNet (Xu et al., 2021) | 26.43 | 0.8016 | 0.1966 | 37.04 | 0.9435 | 0.0958 | 35.60 | 0.9380 | 0.1246 | 33.51 | 0.9159 | 0.1638 | 30.59 | 0.8760 | 0.1744 | 14.97 | 12.26 |
| VideoINR (Chen et al., 2022) | 25.61 | 0.7709 | 0.2489 | – | – | – | – | – | – | – | – | – | 29.92 | 0.8746 | 0.1847 | 15.20 | 11.31 |
| RSTT-L (Geng et al., 2022) | 26.43 | 0.7994 | 0.1973 | 36.80 | 0.9403 | 0.1007 | 35.66 | 0.9381 | 0.1219 | 33.50 | 0.9147 | 0.1634 | – | – | – | 14.14 | 7.67 |
| MoTIF (Chen et al., 2023a) | 25.79 | 0.7745 | – | – | – | – | – | – | – | – | – | – | 29.82 | 0.8750 | – | – | 12.55 |
| SAFA (Huang et al., 2024) | 25.98 | 0.7807 | – | 37.38 | 0.9457 | 0.0949 | 35.65 | 0.9384 | 0.1228 | 33.83 | 0.9184 | 0.1630 | 30.13 | 0.8782 | – | – | 5.00 |
| STDAN (Wang et al., 2023) | 26.28 | 0.8041 | – | 37.10 | 0.9437 | – | 35.70 | 0.9387 | – | 33.66 | 0.9176 | – | – | – | – | 13.80 | 8.29 |
| Hu, Jiang, Nie et al. (2023) | 26.43 | 0.8110 | – | 37.12 | 0.9460 | – | 35.63 | 0.9400 | – | 33.49 | 0.9180 | – | – | – | – | – | 16.40 |
| STINet(Ours) | 26.79 | 0.8049 | 0.1942 | 37.43 | 0.9465 | 0.0943 | 35.87 | 0.9390 | 0.1222 | 34.04 | 0.9180 | 0.1622 | 30.75 | 0.8769 | 0.1738 | 15.33 | 10.89 |
| STINet w/o HR | 26.34 | 0.8025 | 0.1974 | 36.96 | 0.9385 | 0.1016 | 35.39 | 0.9361 | 0.1331 | 33.55 | 0.9160 | 0.1682 | 30.44 | 0.8720 | 0.1772 | 17.17 | – |
| STINet w/o LR | 26.43 | 0.8027 | 0.1985 | 37.10 | 0.9440 | 0.0994 | 35.41 | 0.9337 | 0.1306 | 33.68 | 0.9164 | 0.1678 | 30.58 | 0.8736 | 0.1754 | 15.54 | – |
| STINet w/o ST-LR | 25.41 | 0.7477 | 0.2230 | 35.97 | 0.9344 | 0.1112 | 34.28 | 0.9240 | 0.1527 | 32.70 | 0.9016 | 0.1845 | 29.10 | 0.8248 | 0.2086 | 16.21 | – |
| STINet w/o ST-GR | 25.94 | 0.7785 | 0.2196 | 36.70 | 0.9400 | 0.1037 | 35.11 | 0.9348 | 0.1364 | 33.24 | 0.9130 | 0.1715 | 29.69 | 0.8365 | 0.1950 | 16.46 | – |
| STINet w/o MCL | 26.53 | 0.8037 | 0.1957 | 37.11 | 0.9443 | 0.0959 | 35.70 | 0.9362 | 0.1255 | 33.82 | 0.9165 | 0.1639 | 30.60 | 0.8743 | 0.1746 | 14.80 | – |

**Evaluation protocol.** Like in Geng et al. (2022), Xiang et al. (2020), Xu et al. (2021), the Peak Signal-to-Noise Ratio (PSNR), Structural Similarity Index (SSIM) (Wang, Bovik, Sheikh, & Simoncelli, 2004), and Learned Perceptual Image Patch Similarity (LPIPS) (Zhang, Isola, Efros, Shechtman, & Wang, 2018) metrics are leveraged to evaluate the performance. PSNR measures pixel-level reconstruction errors between the prediction and ground truth. SSIM compares the luminance, contrast and structure differences in the frame-level. LPIPS measures the perceptual similarity between two images by calculating the Euclidean distance of their deep features. By default, we perform single frame interpolation as per most comparable methods (Geng et al., 2022; Haris et al., 2020; Shi et al., 2021; Xiang et al., 2020; Xu et al., 2021). Results of multi-frame interpolation are also provided.

## 4.3. Comparison to state of the art

The state of the art can be divided into two-stage and one-stage approaches. For two-stage approaches, we leverage the combination of latest VFI (SepConv (Niklaus et al., 2017), DAIN (Bao et al., 2019), BMBC (Park et al., 2020), ABME (Park, Lee, & Kim, 2021)) and VSR (Bicubic, RCAN (Zhang, Li et al., 2018), RBPN (Haris et al., 2019), EDVR (Wang et al., 2019), ETDM (Isobe et al., 2022), TTVSR (Liu et al., 2022)) models to achieve STVSR. For one-stage approaches, Zooming Slow-Mo (Xiang et al., 2020), STARnet (Haris et al., 2020), DAVSR (Cao et al., 2022), TMNet (Xu et al., 2021), VideoINR (Chen et al., 2022) RSTT-L (Geng et al., 2022), MoTIF (Chen et al., 2023a) SAFA (Huang et al., 2024), STDAN (Wang et al., 2023) and Hu, Jiang, Nie et al. (2023) are compared.

Table 1 shows the results. For two-stage approaches, having a combination of more recent VFI and VSR models can produce better results than older ones: *e.g.* ABME + ETDM performs the best. Two-stage approaches treat spatial and temporal information separately while ignoring their correlation, their performance is in general inferior to one-stage approaches. Our STINet significantly outperforms the best performing two-stage approach ABME+ETDM by 0.41 dB, 1.19 dB, 0.79 dB, 0.76 dB, 0.21dB on PSNR over five test sets. With respect to the comparison to one-stage approaches, STINet also clearly outperforms most of them on both PSNR and SSIM metrics. For example, on the Vid4 dataset, STINet improves the very recent work

(Geng et al., 2022) by +0.36 dB on PSNR and +0.0055 on SSIM. Moreover, STINet exhibits the best LPIPS among all methods under comparison on four of five test sets, and is only marginally surpassed by RSTT-L on the Vimeo-Medium dataset

We offer visualization results in Fig. 6. It is evident that one-stage approaches generally yield images with higher qualities and finer details than two-stage counterparts. For instance, in the first row, two-stage approaches such as BMBC[53]+ETDM[55] produce blurring artifacts on the reconstructed letters while one-stage approaches such as TMNet (Xu et al., 2021) can recover their sharp edges. This discrepancy arises from the fact that two-stage approaches separate the spatial and temporal processing in STVSR. Furthermore, we can see that STINet shows more visually appealing results than others, particularly in terms of the edges and textures of objects. For instance, in the second row, its reconstructed steel wire mesh is clearer than others; in particular, almost all methods can only generate blurry wires on the left side of the image because of the fast motion, while STINet produces clearer and more complete wires. Moreover, in the first row, STINet can well recover the letter "H". In the last row, STINet excels in reconstructing clear textures inside the leaf.

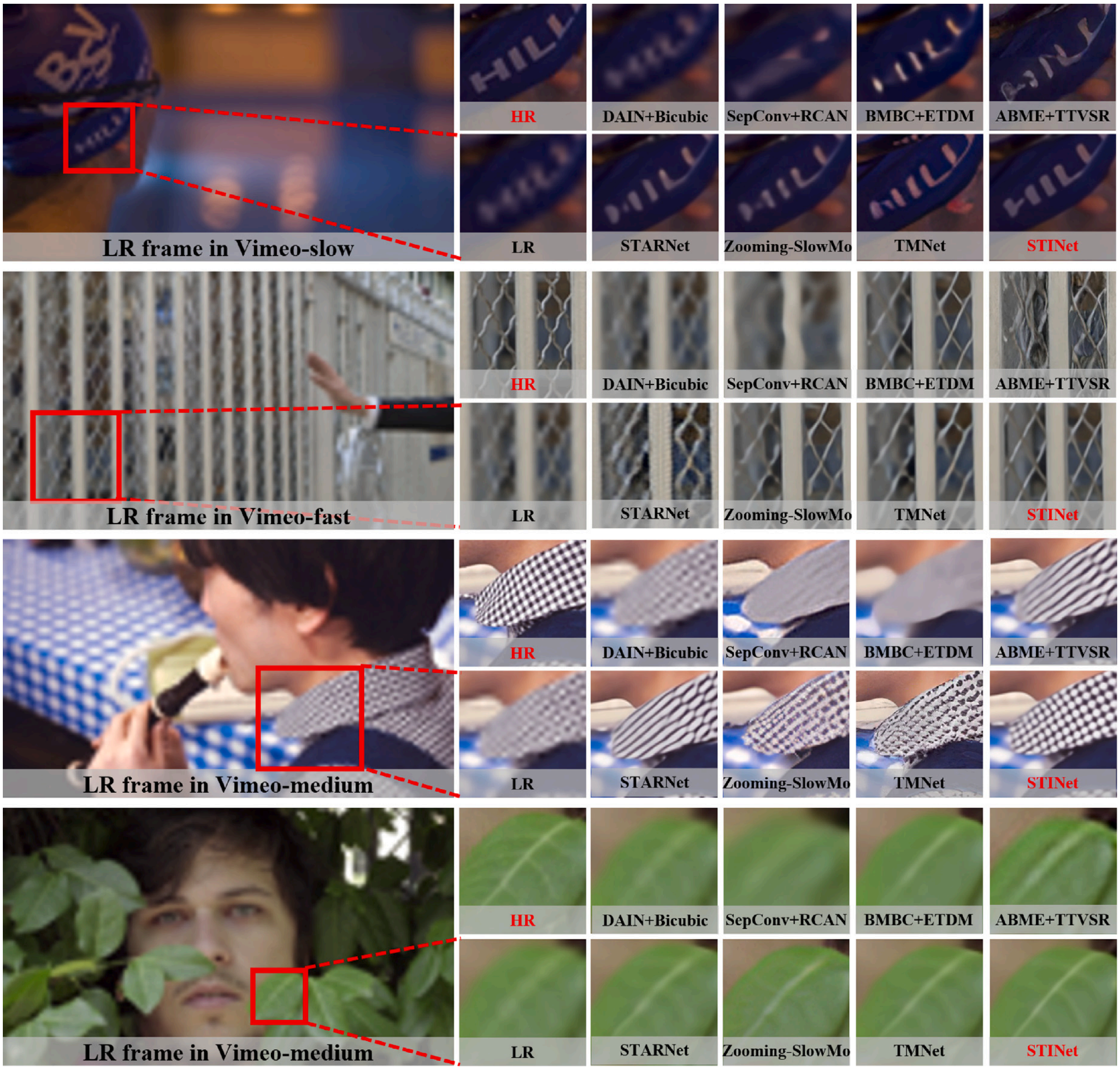

**Fig. 6.** Visual comparison between different STVSR approaches. Our STINet show more visually appealing results than others (zoom in for more details). Ground truth is given as HR.

**The number of parameters and inference speeds.** In terms of computation cost and inference speed, although it seems that STINet has complex network designs, its key components indeed do not involve an extensive number of parameters, ensuring that the network overall efficiency remains comparable to other methods. We have reported the number of parameters and inference speeds on the Vid4 dataset in Table 1, using one Nvidia RTX 3090 GPU. We can observe that the inference speed of our STINet is 15.33fps, surpassing most models. The number of parameters for STINet is 10.89M, which is also comparable to that of state of the art. DAVSR, STDAN and RSTT-L have lighter models, but we clearly outperform them in terms of PSNR, SSIM, LPIPS and speed by a considerable margin.

Moreover, we also make a detailed comparison of the number of parameters and inference speed between each key module of STINet and TMNet (Xu et al., 2021).

Table 2 illustrates that our ST-FI and ST-GR modules have saved a lot of parameters compared to TMNet. Specifically, in the feature interpolation phase, TMNet applies multiple DCN layers to align adjacent frames to intermediate frames in forward and backward directions, while our ST-FI only utilizes several convolutional layers for feature alignment based on the adapted optical flows. In the global refinement phase, our ST-GR has fewer parameters than BDConvLSTM in TMNet but with better performance. The reason is that the graph topology with our designed edge attributes is particularly suitable to exchange the spatial–temporal information among feature sequences. These results demonstrate the effectiveness and superiority of the key components of STINet.

**Multi-frame interpolation** STINet supports multi-frame reconstruction between two adjacent LR frames. Table 3 presents the results of reconstructing 3, 5, 7, 9 intermediate frames. Notice that the Vimeo-90K *septuplet test* set only contains video samples with 7 consecutive frames, which means it does not support interpolating more than 5 intermediate frames given two adjacent frames. Therefore, for 7-frame and 9-frame interpolation, we only report the model performance on Vid4 and Adobe240 datasets. Moreover, among the state of the art, TMNet (Xu et al., 2021) supports multi-frame interpolation, so we compare to it. We can notice that STINet yields superior performance to TMNet which demonstrates its strong capability to broadcast low-frame-rate LR video sequences to high-frame-rate HR videos.

**Table 2**
Comparisons of the number of parameters and inference speed between the key modules of TMNet and STINet.

| Method | Feature interpolation phase | | | Local feature refinement phase | | | Global feature refinement phase | | |
|---|---|---|---|---|---|---|---|---|---|
| | Module | Param(M↓) | Speed(fps↑) | Module | Param(M↓) | Speed(fps↑) | Module | Param(M↓) | Speed(fps↑) |
| TMNet | CFI | 6.62 | 27.70 | LFC | **3.09** | **58.82** | BDConvLSTM | 1.54 | 58.49 |
| STINet | ST-FI | **5.04** | **35.19** | ST-LR | 5.32 | 35.65 | ST-GR | **0.48** | **86.92** |

**Table 3**
Results on multi-frame interpolation. Method(X) indicates interpolating X frames given two adjacent frames.

| Method | Vid4 | | | Vimeo-Fast | | | Vimeo-Medium | | | Vimeo-Slow | | | Adobe240 | | |
|---|---|---|---|---|---|---|---|---|---|---|---|---|---|---|---|
| | PSNR↑ | SSIM↑ | LPIPS↓ | PSNR↑ | SSIM↑ | LPIPS↓ | PSNR↑ | SSIM↑ | LPIPS↓ | PSNR↑ | SSIM↑ | LPIPS↓ | PSNR↑ | SSIM↑ | LPIPS↓ |
| TMNet(3) | 25.45 | 0.7488 | 0.2361 | 36.28 | 0.9320 | 0.1075 | 34.52 | 0.9273 | **0.1528** | 32.75 | 0.9059 | 0.1840 | 29.44 | 0.8315 | 0.2016 |
| **STINet(3)** | **25.55** | **0.7579** | **0.2359** | **36.41** | **0.9335** | **0.1041** | **34.66** | **0.9280** | 0.1530 | **32.91** | **0.9086** | **0.1792** | **29.57** | **0.8323** | **0.1925** |
| TMNet(5) | 23.72 | 0.6445 | 0.3495 | 34.74 | 0.9258 | 0.1396 | 32.70 | 0.9054 | 0.1869 | 31.26 | 0.8720 | **0.1967** | 28.03 | 0.8162 | 0.2227 |
| **STINet(5)** | **23.88** | **0.6557** | **0.3467** | **34.82** | **0.9277** | **0.1377** | **32.97** | **0.9095** | **0.1855** | **31.40** | **0.8786** | 0.1973 | **28.30** | **0.8176** | **0.2178** |
| TMNet(7) | 22.57 | 0.5948 | 0.3714 | – | – | – | – | – | – | – | – | – | 26.85 | 0.7488 | 0.2400 |
| **STINet(7)** | **22.91** | **0.6060** | **0.3673** | – | – | – | – | – | – | – | – | – | **27.19** | **0.7520** | **0.2382** |
| TMNet(9) | 21.95 | 0.5573 | 0.3920 | – | – | – | – | – | – | – | – | – | 25.70 | 0.7349 | 0.2541 |
| **STINet(9)** | **21.98** | **0.5581** | **0.3909** | – | – | – | – | – | – | – | – | – | **25.78** | **0.7350** | **0.2534** |

**Table 4**
Ablation study on the adapted optical flows.

| Method | Vid4 | | Vimeo-Fast | | Vimeo-Medium | | Vimeo-Slow | | Adobe240 | |
|---|---|---|---|---|---|---|---|---|---|---|
| | PSNR↑ | SSIM↑ | PSNR↑ | SSIM↑ | PSNR↑ | SSIM↑ | PSNR↑ | SSIM↑ | PSNR↑ | SSIM↑ |
| STINet w/o optical flows | 26.15 | 0.7950 | 36.71 | 0.9403 | 35.30 | 0.9354 | 33.59 | 0.9157 | 30.00 | 0.8666 |
| STINet w/o flow adaption | 26.62 | 0.8031 | 37.30 | 0.9448 | 35.79 | 0.9381 | 34.00 | 0.9178 | 30.59 | 0.8748 |
| **STINet** | **26.79** | **0.8049** | **37.43** | **0.9465** | **35.87** | **0.9390** | **34.04** | **0.9180** | **30.75** | **0.8769** |

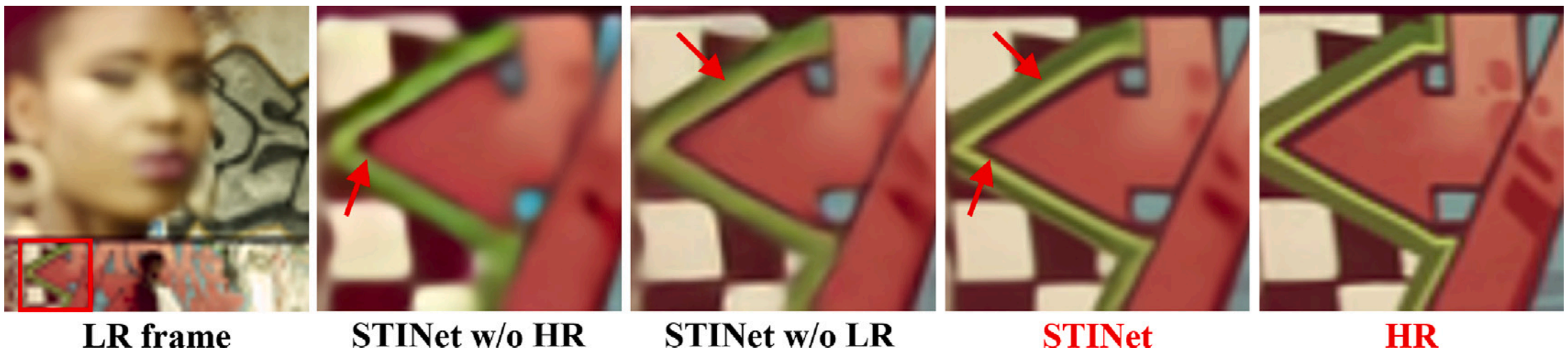

**Fig. 7.** Visual results of the ablation study regarding LR and HR features. Ground truth is given as HR.

### *4.4. Ablation study*

Table 1 and Fig. 7 present the basic ablation study on LR and HR features. We use *STINet w/o HR* to indicate that we only initialize and interpolate LR features in the ST-FI module. The subsequent ST-LR and ST-GR modules are hence degraded to only work on LR features without LR-HR interaction. We observe a clear decrease on PSNR and SSIM from STINet to *STINet w/o HR*, and the latter one reconstructs the red arrow with blurred structures and edges in Fig. 7. Similar observations can be found on *STINet w/o LR* where we only use HR features in the main body of the network. It fails to reconstruct a distinct boundary between the dark-green and light-green stripes. The performance decrease of these two variants justify our idea of performing comprehensive interactions among features of different resolutions to improve STVSR. Below we detail the ablation study of each module of STINet.

**Spatial–temporal frame interpolation module (ST-FI).**

*Adapted optical flows:* Referring to Eq. (1), in order to interpolate the intermediate frame feature, we first propose to adapt the original optical flows between two given frames to that from the intermediate frame to given frames. We evaluate the effectiveness of our adapted optical flows in Table 4: *STINet w/o optical flows* denotes the ST-FI module without using optical flows; *STINet w/o flow adaption* denotes the ST-FI module using the original optical flows between given frames without adaption, similar to Haris et al. (2020). It can be observed that without using optical flows, the PSNR of STINet significantly decreases by 0.64 dB, 0.72 dB, 0.57 dB, 0.45 dB, 0.75 dB on five test sets, respectively. Moreover, when using the optical flows yet without flow adaption, the performance of *STINet w/o flow adaption* is better than that of *STINet w/o optical flows* but is inferior to STINet. The adapted flows are more accurate to describe motions for the intermediate frame interpolation.

*Shareable blocks:* next, we evaluate the importance of having shareable residual blocks in ST-FI. Table 5 presents the results using separate residual blocks in the LR and HR interpolation branches (three residual blocks each), *i.e. ST-FI w/o share*. We can observe that the PSNR decreases by 0.29 dB, 0.32 dB, 0.20 dB, 0.18 dB, 0.32 dB on five test sets, respectively. This demonstrates that the shareable blocks between LR and HR features are beneficial to encode mutual and complementary information between them and thus help improve the interpolation performance.

**Spatial–temporal local refinement module (ST-LR).** We first present the basic ablation study on the proposed ST-LR. We use *STINet w/o ST-LR* to denote the variant in which we remove ST-LR from STINet. As shown in Table 1 and Fig. 8, the performance clearly declines on all datasets for this variant, and it exhibits poor visual performance in the two samples of Fig. 8. The structures and edges of the letters are severely deteriorated in the reconstructed images. The initially interpolated features from the ST-FI module are coarse in reflecting spatial–temporal contexts, ST-LR plays a crucial role in refining their details and eliminating their errors by utilizing the motion cues from temporal neighbors of features. Hence, without ST-LR, it becomes difficult for the network to restore fine details.

**Table 5**
Ablation study on the proposed ST-FI, ST-LR and ST-GR.

| Method | Vid4 | | Vimeo-Fast | | Vimeo-Medium | | Vimeo-Slow | | Adobe240 | |
|---|---|---|---|---|---|---|---|---|---|---|
| | PSNR↑ | SSIM↑ | PSNR↑ | SSIM↑ | PSNR↑ | SSIM↑ | PSNR↑ | SSIM↑ | PSNR↑ | SSIM↑ |
| ST-FI w/o share | 26.50 | 0.8015 | 37.11 | 0.9432 | 35.67 | 0.9355 | 33.86 | 0.9160 | 30.43 | 0.8727 |
| ST-LR w/o Inet | 26.65 | 0.8033 | 37.30 | 0.9451 | 35.71 | 0.9372 | 33.96 | 0.9176 | 30.54 | 0.8738 |
| ST-LR w/o $s^H$ | 26.63 | 0.8030 | 37.31 | 0.9453 | 35.68 | 0.9366 | 34.00 | 0.9170 | 30.57 | 0.8741 |
| ST-LR w/o $s^L$ | 26.69 | 0.8040 | 37.36 | 0.9450 | 35.74 | 0.9370 | 33.96 | 0.9173 | 30.59 | 0.8744 |
| ST-GR w/o $E^F$ | 26.70 | 0.8038 | 37.37 | 0.9450 | 35.83 | 0.9365 | 33.91 | 0.9162 | 30.53 | 0.8741 |
| ST-GR w/o $E^P$ | 26.72 | 0.8040 | 37.36 | 0.9457 | 35.82 | 0.9371 | 33.90 | 0.9167 | 30.56 | 0.8740 |
| ST-GR w/o $E^T$ | 26.74 | 0.8047 | 37.38 | 0.9455 | 35.83 | 0.9374 | 33.94 | 0.9166 | 30.68 | 0.8750 |
| ST-GR→BDCL | 26.71 | 0.8045 | 37.40 | 0.9459 | 35.77 | 0.9380 | 33.93 | 0.9179 | 30.58 | 0.8738 |
| **STINet** | **26.79** | **0.8049** | **37.43** | **0.9465** | **35.87** | **0.9390** | **34.04** | **0.9180** | **30.75** | **0.8769** |

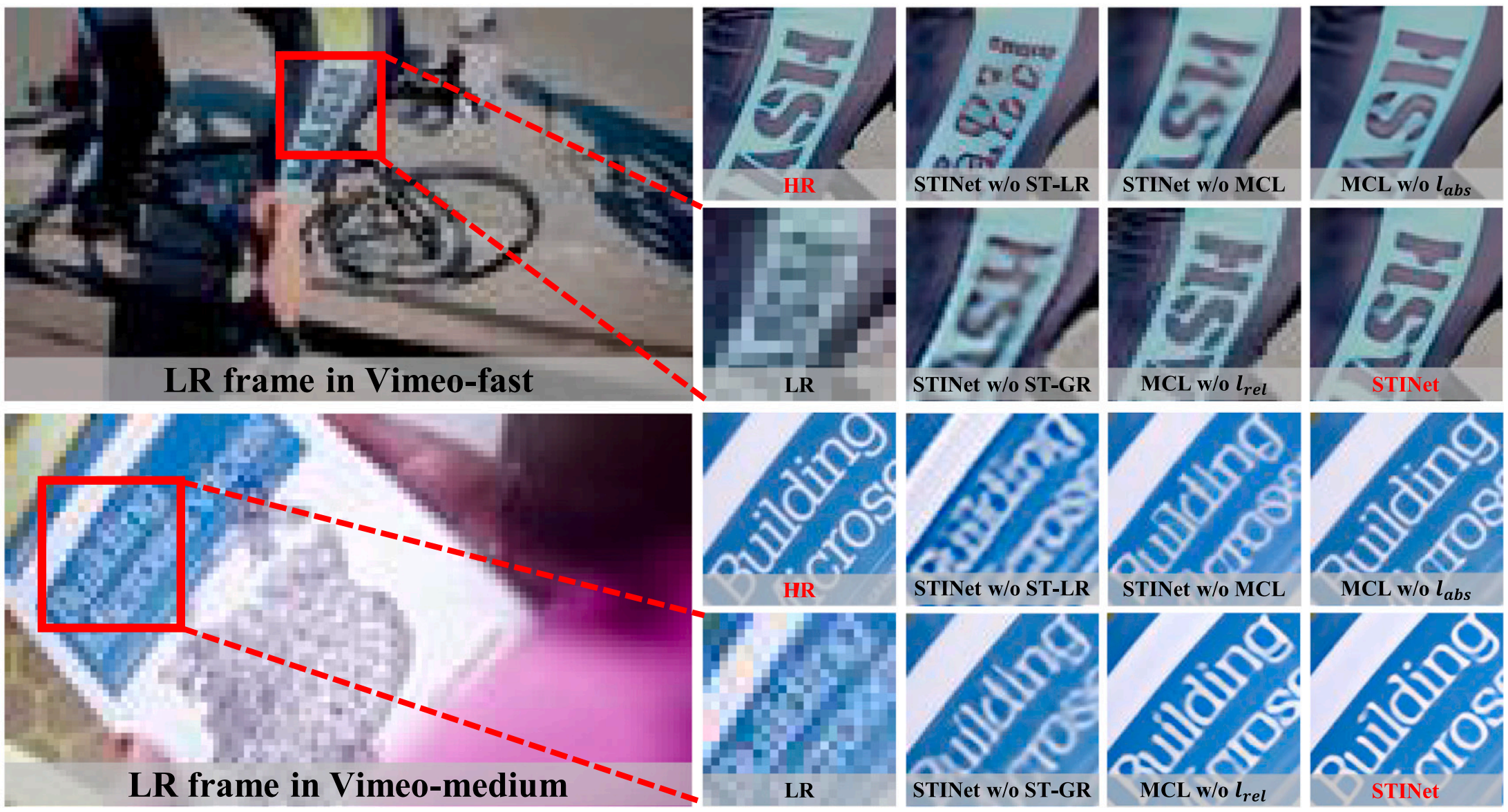

**Fig. 8.** Visual results of the ablation study regarding the ST-LR, ST-GR and MCL. Ground truth is given as HR.

Moreover, in ST-LR, we learn two sets of offsets that have the same temporal stamp but different spatial resolutions and we introduce an interaction subnet (Inet) to enable their interaction. First, if we only predict one set of offsets $s^L$ (or $s^H$) from LR (or HR) features and use it for the local feature alignment in both LR and HR branches, the performance will decrease. We denote the variant by *ST-LR w/o* $s^H$(or *ST-LR w/o* $s^L$) in Table 5. We can observe that both $s^L$ and $s^H$ contribute to ST-LR; *e.g.* in Vid4, our STINet shows a decrease of PNSR by −0.10 dB and −0.16 dB without $s^L$ and $s^H$, respectively.

Next, we verify the benefits of designing Inet in the ST-LR module. In Table 5, *ST-LR w/o Inet* denotes ST-LR without Inet such that the learned offsets are directly fed into their respective DCN layers. The performance decreases substantially on all datasets after removing the Inet.

Above results show that the corresponding offsets learned from LR and HR features contain complementary information, and it is essential to interact them for better local feature refinement.

**Spatial–temporal global refinement module (ST-GR).** First, we explore the contribution of ST-GR module to the overall model performance. In Table 1 and Fig. 8, we use *STINet w/o ST-GR* to signify the variant where the ST-GR module is excluded from STINet. We can observe that this variant has poor performance in Table 1, and it displays motion artifacts on the reconstructed letters, such as blurry "S" and "H" in the first sample of Fig. 8. ST-GR captures spatial–temporal information for global feature refinement. The absence of it would deprive STINet from modeling motions over the whole video, making it challenging to maintain long-term temporal consistency.

Moreover, we investigate the effectiveness of the designed edge attributes in the graph of the ST-GR module. The results are shown in Table 5. We use *ST-GR w/o* $E^F$, *ST-GR w/o* $E^P$ and *ST-GR w/o* $E^T$ to denote the graph without edge attribute $E^F$, $E^P$ and $E^T$ (Section 3.4). One can clearly see performance decrease by these variants compared to the original ST-GR (STINet). The three edge attributes play an important role in ST-GR: edges between visually similar nodes are assigned with bigger $E^F$ and $E^P$ to capture their mutual information for recovering object details; edges between temporally close nodes are assigned with bigger $E^T$ to exchange more information to reinforce motion continuity. We also show visual results of these variants in Fig. 9. For example, in the second sample of a bird waving its wings, we can see these variants struggle to reconstruct clear and accurate textures of the wings. In contrast, STINet synthesizes the most reliable and plausible results. These results further demonstrate that enhancing the exchange of complementary spatial–temporal information between visually similar or temporally close nodes is beneficial to generating more accurate details in reconstructed frames.

Next, we compare our ST-GR with the widely used BDConvLSTM layer in previous works (Xiang et al., 2020; Xu et al., 2021) for global feature refinement. We denote by *ST-GR→BDCL* replacing ST-GR with BDConvLSTM in STINet. Again, the results are inferior to the original STINet. This suggests the superiority of using our ST-GR over BDConvLSTM to globally aggregate spatial–temporal contexts for better feature refinement.

**Motion consistency loss (MCL).** We first use *STINet w/o MCL* to denote the variant of STINet where MCL is not utilized for network optimization, and report its performance in Table 1 and Fig. 8. We can observe that its PSNR metric is 0.26 dB, 0.32 dB, 0.17 dB, 0.22 dB, 0.15 dB lower than that of the original STINet on five datasets, respectively. Moreover, it shows slight blurry effects on words, *e.g.* , "building" in the second sample of Fig. 8. MCL computes predicted optical flows across reconstructed frames, and then enforces the predictions to conform to both the ground truth and the motion continuity between frames. The absence of MCL may result in non-smooth object motions and introduce blurring effects in frames.

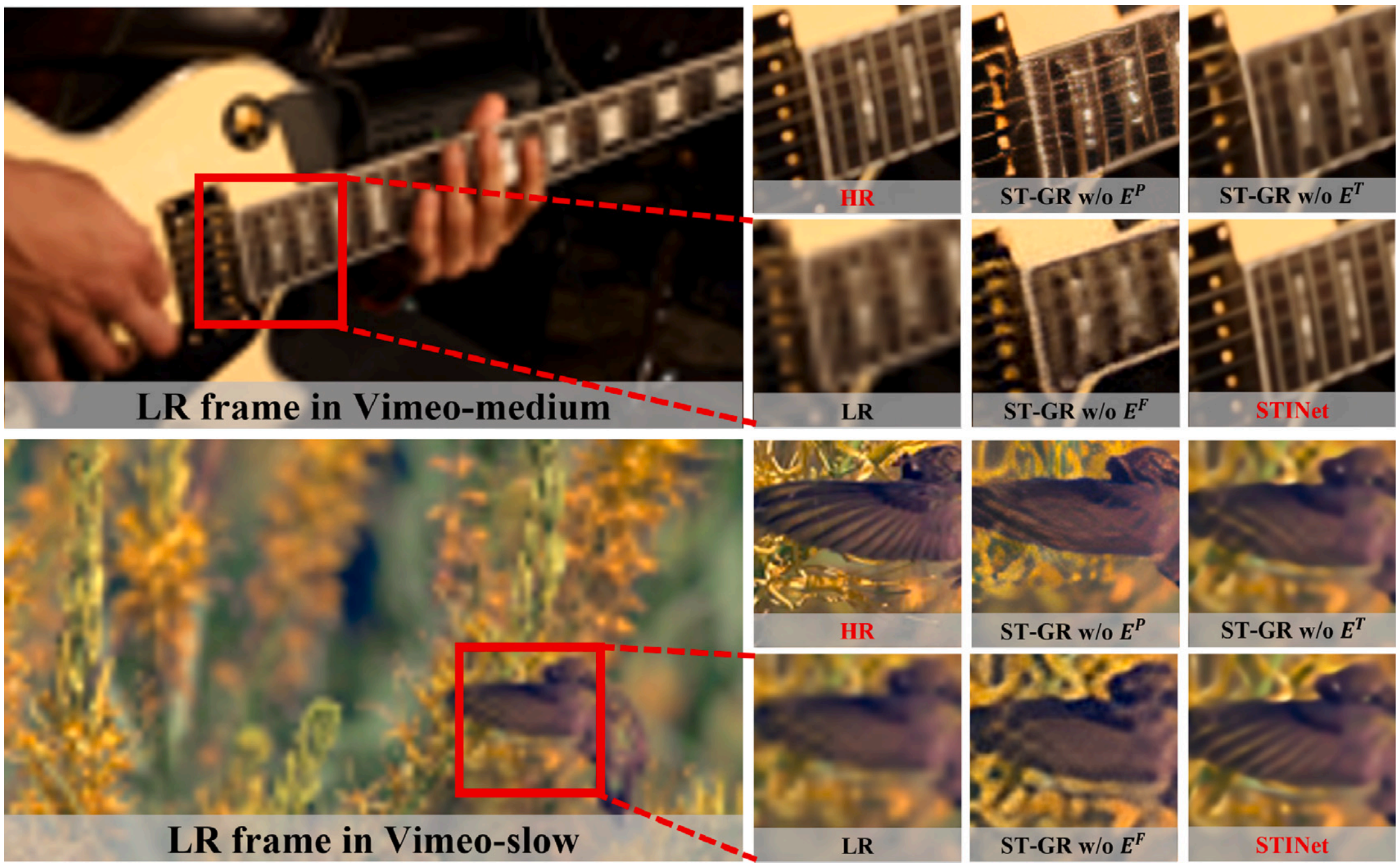

**Fig. 9.** Visual results of the ablation study regarding the three edge attributes in the graph of the ST-GR module. Ground truth is given as HR.

**Table 6**
Ablation study on the proposed MCL.

| Method | Vid4 | | Vimeo-Fast | | Vimeo-Medium | | Vimeo-Slow | | Adobe240 | |
|---|---|---|---|---|---|---|---|---|---|---|
| | PSNR↑ | SSIM↑ | PSNR↑ | SSIM↑ | PSNR↑ | SSIM↑ | PSNR↑ | SSIM↑ | PSNR↑ | SSIM↑ |
| MCL w/o $l_{abs}$ | 26.72 | 0.8044 | 37.28 | 0.9444 | 35.77 | 0.9376 | 33.93 | 0.9170 | 30.64 | 0.8751 |
| MCL w/o $l_{rel}$ | 26.63 | 0.8039 | 37.17 | 0.9443 | 35.72 | 0.9366 | 33.88 | 0.9167 | 30.60 | 0.8747 |
| MCL w/ $l_{abs-AJ}$ | 26.74 | 0.8047 | 37.40 | 0.9465 | 35.86 | 0.9387 | 34.01 | 0.9171 | 30.68 | 0.8759 |
| MCL w/ $l_{abs-L1}$ | 26.73 | 0.8045 | 37.34 | 0.9450 | 35.82 | 0.9382 | 33.96 | 0.9175 | 30.68 | 0.8760 |
| MCL w/ $l_{rel-SC}$ | 26.70 | 0.8038 | 37.33 | 0.9444 | 35.80 | 0.9381 | 33.92 | 0.9169 | 30.62 | 0.8753 |
| **MCL (STINet)** | **26.79** | **0.8049** | **37.43** | **0.9465** | **35.87** | **0.9390** | **34.04** | **0.9180** | **30.75** | **0.8769** |

Moreover, MCL consists of two loss terms, the absolute term $l_{abs}$ (Eq. (2)) and the relative term $l_{rel}$ (Eq. (3)). To validate their effectiveness, we remove either one and re-train the model. Fig. 8 shows the qualitative results: some motion artifacts are observed on the reconstructed letters due to the absence of either $l_{abs}$ or $l_{rel}$ in MCL,*e.g.* the blurry "o" and "s" in the second sample. Table 6 shows the quantitative results: the PSNR and SSIM decline in all datasets for *MCL w/o* $l_{abs}$ or $l_{rel}$. Without them, the motions of objects may not be smooth and blurring artifacts may be produced in frames. This is especially the case for the Vimeo-Fast dataset with fast motions. Specifically, we notice that the decline of *MCL w/o* $l_{rel}$ is bigger than that of *MCL w/o* $l_{abs}$: for instance, the PSNR decline of the former is 0.16 dB while that of the latter is 0.07 dB in Vid4. This shows that $l_{rel}$ is more important than $l_{abs}$ in the motion consistency loss.

Next, we present some variants corresponding to $l_{abs}$ and $l_{rel}$, respectively. For $l_{abs}$, referring to Section 3.6, we distinguish it from the flow losses in Huang et al. (2021), Zhou et al. (2021): (1) (Huang et al., 2021; Zhou et al., 2021) only supervise flows between adjacent frames; if we do this, denoted by *MCL w/* $l_{abs-AJ}$, the performance will be reduced from the original MCL, *e.g.* in Adobe240, −0.07 dB on PSNR and −0.0010 on SSIM. (2) (Huang et al., 2021; Zhou et al., 2021) use L1 loss; if we also use L1 loss, denoted by *MCL w/* $l_{abs-L1}$, the performance will also be reduced, *e.g.* in Vid4, -0.06 dB on PSNR and −0.0004 on SSIM.

For $l_{rel}$, we design it as a weak constraint rather than a strong one to avoid overfitting. If we instead optimize it upon the exact difference between $O^{pred}_{t-1\to t} - O^{pred}_{t-1\to t+1}$ and $O^{gt}_{t-1\to t} - O^{gt}_{t-1\to t+1}$, denoted by *MCL w/* $l_{rel-SC}$ in Table 6, the performance will decrease compared to the original MCL: *e.g.* the PSNR and SSIM are −0.09 dB and −0.0011 on Vid4.

### *4.5. Parameter variation*

We vary the loss weights ($\lambda_1$ and $\lambda_2$) for the perceptual loss and the proposed MCL in Eq. (4) and report the performance on the Vid4 dataset in Fig. 10. We can observe that the best performance occurs when $\lambda_1 = 0.1$ and $\lambda_2 = 0.1$, which is our default setting.

### *4.6. Method generalizability*

Our proposed components are not only effective in STINet but can also generalize to other methods. We first take the state-of-the-art method TMNet (Xu et al., 2021) as a baseline to add the proposed components in our STINet. The results are shown in Table 7: (1) we add HR features to the TMNet (denoted as *TMNet w/ HR*) such that the LR and HR features are respectively refined via two separate branches and finally fused for video reconstruction. Sizeable improvement can be observed upon the original TMNet: *i.e.* in Vid4, we obtain +0.24 dB on PSNR and +0.0017 on SSIM. This validates our idea of exploiting both LR and HR features in the STVSR task. Our STINet further improves upon this idea by introducing spatial–temporal interaction between LR and HR features. (2) We add MCL to TMNet (denoted by *TMNet w/ MCL*) and also observe sizeable improvement: *e.g.* in Vid4, 26.64 dB on PSNR and 0.8029 on SSIM, which demonstrates the advantage of MCL. (3) Moreover, we carefully evaluate the generalizability of the proposed ST-GR since it differs substantially with the widely used BDConvLSTM layer for global feature refinement. Specifically, we incorporate ST-GR into the architectures of TMNet (Xu et al., 2021), Zooming Slow-Mo (Xiang et al., 2020) and STARnet (Haris et al., 2020). However, TMNet and Zooming Slow-Mo do not encode HR features in their pipelines. Therefore, we first degrade the graph of ST-GR by removing its nodes corresponding to HR features. Then we replace the BDConvLSTM layer in TMNet and Zooming Slow-Mo with this degraded version of ST-GR for feature refinement. These variants are denoted as *TMNet w/ ST-GR* and *Zooming Slow-Mo w/ ST-GR*. As for STARnet, we directly equip it with the original ST-GR (denoted by *STARnet w/ ST-GR*). Table 8 illustrates performance improvements for these three methods: *e.g.* , in Adobe240, adding ST-GR to TMNet, Zooming Slow-Mo, and STARnet leads to +0.06 dB, +0.05 dB, and +0.46 dB on PSNR respectively. These results demonstrate the good generalizability of ST-GR for the STVSR task.

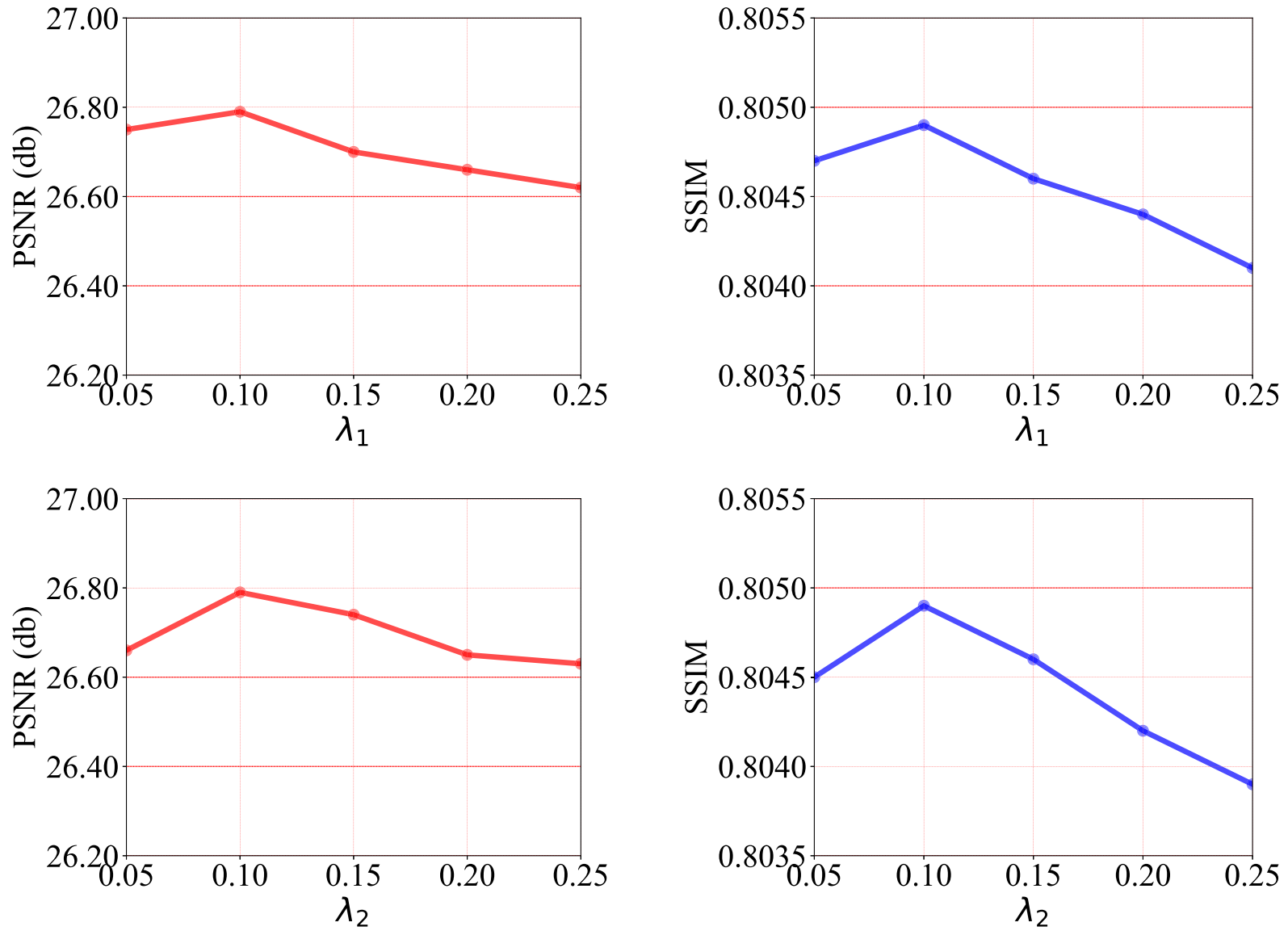

**Fig. 10.** The effect of different loss weights ($\lambda_1$ and $\lambda_2$) in our loss function. The results are reported on the Vid4 dataset.

**Table 7**
Results of adding key components of STINet to TMNet (Xu et al., 2021).

| Method | Vid4 | | Vimeo-Fast | | Vimeo-Medium | | Vimeo-Slow | | Adobe240 | |
|---|---|---|---|---|---|---|---|---|---|---|
| | PSNR↑ | SSIM↑ | PSNR↑ | SSIM↑ | PSNR↑ | SSIM↑ | PSNR↑ | SSIM↑ | PSNR↑ | SSIM↑ |
| TMNet | 26.43 | 0.8016 | 37.04 | 0.9435 | 35.60 | 0.9380 | 33.51 | 0.9159 | 30.59 | 0.8760 |
| TMNet w/ HR | 26.67 | 0.8033 | 37.29 | 0.9450 | 35.81 | 0.9382 | 33.77 | 0.9168 | 30.70 | 0.8765 |
| TMNet w/ MCL | 26.64 | 0.8029 | 37.20 | 0.9439 | 35.69 | 0.9385 | 33.59 | 0.9163 | 30.68 | 0.8762 |

**Table 8**
Generalizability validation on the ST-GR module.

| Method | Vid4 | | Vimeo-Fast | | Vimeo-Medium | | Vimeo-Slow | | Adobe240 | |
|---|---|---|---|---|---|---|---|---|---|---|
| | PSNR↑ | SSIM↑ | PSNR↑ | SSIM↑ | PSNR↑ | SSIM↑ | PSNR↑ | SSIM↑ | PSNR↑ | SSIM↑ |
| TMNet | 26.43 | 0.8016 | 37.04 | 0.9435 | 35.60 | 0.9380 | 33.51 | 0.9159 | 30.59 | 0.8760 |
| TMNet w/ ST-GR | **26.50** | **0.8021** | **37.07** | **0.9435** | **35.70** | **0.9382** | **33.59** | **0.9163** | **30.65** | **0.8761** |
| Zooming Slow-Mo | 26.31 | **0.7976** | 36.81 | 0.9415 | 35.41 | **0.9361** | 33.36 | 0.9138 | 30.34 | 0.8713 |
| Zooming Slow-Mo w/ ST-GR | **26.38** | 0.7974 | **36.89** | **0.9423** | **35.54** | 0.9360 | **33.44** | **0.9146** | **30.39** | **0.8720** |
| STARnet | 26.06 | 0.8046 | 36.19 | 0.9368 | 34.86 | 0.9356 | 33.10 | 0.9164 | 29.92 | 0.8589 |
| STARnet w/ ST-GR | **26.48** | **0.8057** | **37.01** | **0.9379** | **35.33** | **0.9372** | **33.50** | **0.9176** | **30.38** | **0.8634** |
| STINet | 26.79 | 0.8049 | 37.43 | 0.9465 | 35.87 | 0.9390 | 34.04 | 0.9180 | 30.75 | 0.8769 |

## 5. Conclusion

In this paper, we propose a spatial–temporal feature interaction network (STINet) for STVSR. Our main contribution is the facilitation of spatial–temporal information interaction among LR and HR features of different frames. Specifically, we introduce the ST-FI module for LR and HR feature interpolation, and the ST-LR module to capture motion cues between features of each frame and its neighbors for local refinement. Moreover, the ST-GR module is presented to perform spatial–temporal information passing and exchanging over the whole feature sequence for global refinement. Last, the proposed motion consistency loss can reinforce STINet to maintain motion continuity among reconstructed frames. Experiments on three datasets demonstrated the effectiveness and superiority of STINet over state of the art. In our future work, we plan to further improve the inference speed of STINet to achieve real-time STVSR.